\documentclass[letterpaper]{article}
\usepackage{proceed2e}
\usepackage[margin=1in]{geometry}
\usepackage{times}

\usepackage[dvipsnames]{xcolor}
\usepackage[square,numbers,sort]{natbib}

\usepackage{xspace}
\usepackage[disable]{todonotes}
\usepackage{amsmath,amssymb,amsfonts}
\usepackage{amsthm}

\usepackage{subcaption}
\usepackage{bm}
\usepackage{bbm} 
\usepackage[lined, boxed, ruled, commentsnumbered, noend]{algorithm2e}

\usepackage[pdftex,bookmarks=true,bookmarksopen=true,bookmarksnumbered]{hyperref}
\usepackage{soul} 
\hypersetup{
  colorlinks = true,
  citecolor = MidnightBlue,
  linkcolor = BrickRed,
}

\usepackage[
toc,
numberedsection=nolabel,
section=section,
]{glossaries}

\newglossary[slg]{symbolslist}{syi}{syg}{Notations} 
\newglossary{accronymslist}{aci}{acg}{Terms and Abbreviations}

\newglossarystyle{notationlong}{%
  \setglossarystyle{long}
  \setglossarystyle{long}
  \renewenvironment{theglossary}{
    \begin{longtable}{@{}p{\dimexpr 3cm-\tabcolsep}p{0.8\hsize}}}%
    {\end{longtable}}%

}

\newglossarystyle{acronymlong}{%
  \setglossarystyle{long}
    {\end{longtable}}%

}

\newcommand*{\B}[1]{\ifmmode\bm{#1}\else\textbf{#1}\fi}

\newcommand{\policy}{\pi}

\newcommand{\obj}{f}

\newcommand{\region}{{\mathcal{R}}}
\newcommand{\Obs}{X}
\newcommand{\obs}{x}
\newcommand{\bobs}{\bx}

\newcommand{\bObs}{\mathbf{X}}

\newcommand{\decision}{d} 
\newcommand{\Actionset}{\cD}
\newcommand{\Testset}{\cT}
\newcommand{\test}{t}
\newcommand{\numtest}{n}
\newcommand{\numrc}{m}
\newcommand{\condprob}{\theta}
\newcommand{\bcondprob}{\btheta}

\newcommand{\Hiddenvar}{Y}
\newcommand{\hiddenvar}{y}
\newcommand{\hiddenVarDom}{\cY}
\newcommand{\obsDom}{\{0,1\}}

\newcommand{\valueof}[1]{\VoI(#1)} 
\newcommand{\expctutilof}[1]{U(#1)} 

\newcommand{\played}[2]{\cS(#1,#2)}

\DeclareMathOperator{\cost}{cost}
\DeclareMathOperator{\VoI}{VoI}
\DeclareMathOperator{\Regret}{Regret}

\newcommand{\DIRECT}{{\sc DiRECt}\xspace}
\newcommand{\ECT}{{$\textsc{EC}^2$}\xspace}
\newcommand{\VOI}{{\sc VoI}\xspace}
\newcommand{\HEC}{{\sc HEC}\xspace}

\newcommand{\GBS}{{\sc GBS}\xspace}
\newcommand{\DRD}{{\sc DRD}\xspace}

\newcommand{\commentout}[1]{}

\newcommand{\given}{\mid}

\newcommand{\paren} [1] {\ensuremath{ \left( {#1} \right) }}

\renewcommand{\Pr}[1]{\ensuremath{\mathbb{P}\left[#1\right] }}
\newcommand{\expct}[1]{\mathbb{E}\left[#1\right]}
\newcommand{\expctover}[2]{\mathbb{E}_{#1}\!\left[#2\right]}
\newcommand{\PrOver}[2]{\ensuremath{\mathbb{P}_{#1}\!\left[#2\right]}}
\def \argmax {\mathop{\rm arg\,max}}
\def \argmin {\mathop{\rm arg\,min}}

\newcommand{\NonNegativeReals}{\ensuremath{\mathbb{R}_{\ge 0}}}

\newcommand{\hypotheses}[0]{\mathcal{H}}

\newcommand{\hypothesis}[0]{\ensuremath{h}}

\newcommand{\bigO}[1]{\ensuremath{O\paren{#1}}}

\newcommand{\OPT}{\textsf{OPT}}

\newcommand{\cD}{{\mathcal{D}}}
\newcommand{\cS}{{\mathcal{S}}}
\newcommand{\cA}{{\mathcal{A}}}

\newcommand{\cB}{{\mathcal{B}}}

\newcommand{\cY}{{\mathcal{Y}}}
\newcommand{\cH}{{\mathcal{H}}}

\newcommand{\cT}{{\mathcal{T}}}

\newcommand{\bx}{{\mathbf{x}}}

\newcommand{\balpha}{{\B{\alpha}}}
\newcommand{\bbeta}{{\B{\beta}}}
\newcommand{\btheta}{{\B{\theta}}}

\newcommand{\tH}{{\tilde{\mathcal{H}}}}

\newtheorem{theorem}{Theorem}
\newtheorem{corollary}[theorem]{Corollary}

\newtheorem{lemma}[theorem]{Lemma}

\newcommand{\figref}[1]{Fig.~\ref{#1}}

\newcommand{\secref}[1]{\S\ref{#1}}
\newcommand{\thmref}[1]{Theorem~\ref{#1}}
\newcommand{\corref}[1]{Corollary~\ref{#1}}

\newcommand{\lemref}[1]{Lemma~\ref{#1}}
\newcommand{\algref}[1]{Algorithm~\ref{#1}}

\newcommand{\THMREF}[1]{THEOREM~\ref{#1}}
\newcommand{\ALGREF}[1]{ALGORITHM~\ref{#1}}

\newglossaryentry{am:ect}{name=\ECT,
  description={Equivalence Class Edge Cutting},
  type=accronymslist,
  sort=EC2
}

\newglossaryentry{am:ecd}{name=\textsc{ECD},
  description={Equivalence Class Determination},
  type=accronymslist,
  sort=ECD
}

\newglossaryentry{am:nvoi}{name=NVOI-NMU,
  description={Nonmyopic value of information problem for achieving near-maxmial utility},
  type=accronymslist,
  sort=nvoi
}

\newglossaryentry{am:drd}{name=\DRD,
  description={Decision Region Determination},
  type=accronymslist,
  sort=DRD
}

\newglossaryentry{am:voi}{name=\VOI,
  description={Value of Information},
  type=accronymslist,
  sort=VoI
}

\newglossaryentry{am:ig}{name=\textsc{IG},
  description={Information Gain},
  type=accronymslist,
  sort=IG
}

\newglossaryentry{am:us}{name=\textsc{US},
  description={Uncertainty Sampling},
  type=accronymslist,
  sort=US
}

\newglossaryentry{am:cpt}{name=CPT,
  description={Conditional Probability Table},
  type=accronymslist,
  sort=CPT
}

\newglossaryentry{symb:groundset}{name=\ensuremath{\cT},
  description={The ground set of tests},
  type=symbolslist,
  sort=test _set}

\newglossaryentry{symb:costtest}{name=\ensuremath{c(\test)},
  description={Cost of test $\test$},
  type=symbolslist,
  sort=cost of test}

\newglossaryentry{symb:costpolicy}{name=\ensuremath{c(\cA)},
  description={Cost of a set of test $c(\cA):=\sum_{t\in\cA} c(t)$},
  type=symbolslist,
  sort=cost of test,
}

\newglossaryentry{symb:avgc}{name=\ensuremath{\cost_{av}(\policy)},
  description={The expected cost of a policy},
  type=symbolslist,
  sort=cost of policy average case,
}

\newglossaryentry{symb:wcc}{name=\ensuremath{\cost_{wc}(\policy)},
  description={The worst-case cost of a policy},
  type=symbolslist,
  sort=cost of policy worst case,
}

\newglossaryentry{symb:obj}{name=\ensuremath{\obj(\bobs_\cA)},
  description={Reward function of a set of observations $\bobs_\cA$, objective function},
  type=symbolslist,
  sort=c objective function,
}

\newglossaryentry{symb:policy}{name=\ensuremath{\policy},
  description={Policy, i.e., a (partial) mapping from observation vectors to tests},
  type=symbolslist,
  sort=policy,
}

\newglossaryentry{symb:optpolicy}{name=\ensuremath{\policy^*, \OPT},
  description={An optimal policy, either with the lowest cost for achieving some reward, or with the maximal reward under some budget},
  type=symbolslist,
  sort=policy optimal,
}

\newglossaryentry{symb:played}{name=\ensuremath{\played{\policy}{\hypothesis}},
  description={The set of tests and outcomes seen by policy $\policy$, if hypothesis $\hypothesis$ is realized},
  type=symbolslist,
  sort=policy played,
}

\newglossaryentry{symb:test}{name=\ensuremath{\test},
  description={Test},
  type=symbolslist,
  sort=test element,
}

\newglossaryentry{symb:numtest}{name=\ensuremath{\numtest},
  description={Size of the set of tests},
  type=symbolslist,
  sort=number of tests,
}

\newglossaryentry{symb:cacb}{name=\ensuremath{\cA,\cB},
  description={(sub-) sets of tests},
  type=symbolslist,
  sort=test _subset,
}

\newglossaryentry{symb:bobs}{name=\ensuremath{\bobs_\cA,\bobs_\cB},
  description={Observation of the outcomes of tests in $\cA$, $\cB$},
  type=symbolslist,
  sort=test outcomes,
}

\newglossaryentry{symb:Obstest}{name=\ensuremath{\Obs_\test},
  description={Random variable that represents the outcome of test $\test$},
  type=symbolslist,
  sort=test Outcome,
}

\newglossaryentry{symb:obstest}{name=\ensuremath{\obs_\test},
  description={A specific outcome of test $\test$},
  type=symbolslist,
  sort=test outcome,
}

\newglossaryentry{symb:HiddenvarT}{name=\ensuremath{\Hiddenvar},
  description={Random Variable that represents the hidden state/ root-cause},
  type=symbolslist,
  sort=Root cause,
}

\newglossaryentry{symb:hiddenvarT}{name=\ensuremath{\hiddenvar},
  description={Value of hidden state/ root-cause},
  type=symbolslist,
  sort=root cause
}

\newglossaryentry{symb:hiddenVarDom}{name=\ensuremath{\hiddenVarDom},
  description={The domain of the hidden state values},
  type=symbolslist,
  sort=root cause domain,
}

\newglossaryentry{symb:numrc}{name=\ensuremath{\numrc},
  description={Size of the set of the root-causes},
  type=symbolslist,
  sort=number of root-causes,
}

\newglossaryentry{symb:hypothesis}{name=\ensuremath{\hypothesis,\bobs_\cT},
  description={Hypothesis; a specific realization of the outcomes of all tests},
  type=symbolslist,
  sort=hypothesis,
}

\newglossaryentry{symb:Hypothesis}{name=\ensuremath{H,\bObs_\cT},
  description={Random variable that denotes a hypothesis; the state of the world},
  type=symbolslist,
  sort=Hypothesis
}

\newglossaryentry{symb:hypotheses}{name=\ensuremath{\hypotheses},
  description={Set of hypotheses},
  type=symbolslist,
  sort=hypotheses set
}

\newglossaryentry{symb:tH}{name=\ensuremath{\tH},
  description={Samples of hypotheses 
  },
  type=symbolslist,
  sort=hypotheses set samples
}

\newglossaryentry{symb:decision}{name=\ensuremath{\decision},
  description={realized decision},
  type=symbolslist,
  sort=target
}

\newglossaryentry{symb:Region}{name=\ensuremath{\region_\decision},
  description={Decision region indexed by decision $\decision$},
  type=symbolslist,
  sort=target region
}

\newglossaryentry{symb:condprob}{name=\ensuremath{\condprob_{ij}},
  description={Conditional probability $\Pr{\obs_i = 1\given \hiddenvar_j}$},
  type=symbolslist,
  sort=v conditional probability table element
}

\newglossaryentry{symb:regretepoch}{name=\ensuremath{\tilde{\Delta}},
  description={Regret of an episode/epoch/session for online adaptive information acquisition},
  type=symbolslist,
  sort=conditional regret,
}

\newglossaryentry{symb:horizonlen}{name=\ensuremath{\tau},
  description={Length of a session for online learning; horizon of an episode},
  type=symbolslist,
  sort=online tau,
}

\newglossaryentry{symb:indices}{name=\ensuremath{i,j,k,l,\ell},
  description={Indexing variables for tests/ hidden states/ sessions/ iterations etc. },
  type=symbolslist,
  sort=_index,
}

\newglossaryentry{symb:coverage}{name=\ensuremath{Z},
  description={Coverage of the total probability mass achieved by the hypothesis samples},
  type=symbolslist,
  sort=coverage,
}

\newglossaryentry{symb:condgain}{name=\ensuremath{\Delta(t \given \bx_\cA)},
  description={The expected gain of performing test $t$, conditioned $\bx_\cA$},
  type=symbolslist,
  sort=conditional gain,
}

\newglossaryentry{symb:coveragethres}{name=\ensuremath{\eta},
  description={Coverage threshold},
  type=symbolslist,
  sort=coverage threshold,
}

\makeglossaries

\begin{document}
%
\title{Efficient Online Learning for Optimizing Value of Information: \\
  Theory and Application to Interactive Troubleshooting
}


\author{{\bf Yuxin Chen} \\
Caltech \\
\And
{\bf Jean-Michel Renders\qquad\qquad}  \\
Xerox Research Center Europe~~~~\qquad~~~~\\
\And
{\bf Morteza Haghir Chehreghani}   \\
Xerox Research Center Europe \\
\And
{\bf Andreas Krause}   \\
ETH Zurich \\
}



\maketitle

\begin{abstract}
  \looseness -1
  We consider the optimal value of information problem, where the goal is to sequentially select a set of tests with a minimal cost, so that one can efficiently make the best decision based on the observed outcomes. Existing algorithms are either heuristics with no guarantees, or scale poorly (with exponential run time in terms of the number of available tests). Moreover, these methods assume a known distribution over the test outcomes, which is often not the case in practice.

  We propose a sampling-based online learning framework to address the above issues. First, assuming the distribution over hypotheses is known, we propose a \emph{dynamic hypothesis enumeration} strategy, which allows efficient information gathering with strong theoretical guarantees. We show that with sufficient amount of samples, one can identify a near-optimal decision with high probability. Second, when the parameters of the hypotheses distribution are unknown, we propose an algorithm which learns the parameters progressively via posterior sampling in an \emph{online} fashion. We further establish a rigorous bound on the expected regret. We demonstrate the effectiveness of our approach on a real-world interactive troubleshooting application, and show that one can efficiently make high-quality decisions with low cost.

\end{abstract}


\section{INTRODUCTION}

Optimal information gathering for decision making is a central challenge in artificial intelligence. A classical approach for decision making is the decision-theoretic \emph{value of information} (VoI) \cite{howard66voi}, 
where one needs to find an optimal testing policy which achieves the maximal value of information.
Informally, the goal of the optimal policy is to reduce the uncertainty about some \emph{hidden state} of the system in question, by efficiently probing it via a sequence of \emph{tests} and observations, so that one can make the best \emph{decision} with the minimal cost. For example, consider the automated troubleshooting problem, where a customer reaches a contact center and wants to resolve some problem with her cell phone.
To provide a solution, a virtual agent has to ask the customer a few questions regarding the symptoms of the cellphone to diagnose the root-cause.
Here, ``decision'' corresponds to the solution, ``hidden state'' corresponds to the root-cause which we want to learn about, ``tests'' correspond to questions on the symptoms, and the ``hypothesis space'' consists of full realizations of all tests. We want to develop a virtual agent, which can identify the best solution for the customer with a minimal set of questions asked.

\paragraph{Optimization of VoI.}
\looseness -1 The optimal VoI problem has been studied in various contexts, including active learning \cite{dasgupta04,settles.book12}, Bayesian experimental design \cite{chaloner1995bayesian}, policy making \cite{Runge20111214} and probabilistic planning \cite{smallwoodPOMDP,kaelbling_1998_pomdp}, etc. We refer interested readers to \secref{related} for a more detailed review of the related work.
Deriving optimal policies is NP-hard in general \cite{chakaravarthy07decision}; however, under certain conditions some approximation results are known. In particular, if test outcomes are deterministic functions of the hidden state (i.e., noise-free setting), then a simple greedy algorithm, namely \emph{generalized binary search} (GBS), is guaranteed to provide a near-optimal approximation of the optimal policy in terms of the cost \cite{kosaraju99}.
These results have recently been generalized to decision making, where information gathering policies no longer aim to resolve all uncertainty of the hidden state -- but just enough to make the optimal decision. Such problem, known as the \emph{Decision Region Determination} (\DRD) problem \cite{javdani14near},
relates the problem of learning the optimal policy with maximal utility, to the problem that aims at resolving the uncertainty amongst the decisions.
Following this line of work, \citet{javdani14near} and \citet{chen15submodular} propose a principled framework for optimizing the VoI using surrogate objective functions.
The theoretical guarantees of these algorithms rely on the fact that the objective functions exhibit \emph{adaptive submodularity} \cite{golovin2011adaptive}, a natural diminishing returns property that generalizes the classical notion of submodularity to adaptive policies. It follows that a simple greedy policy can provide near-optimal approximation to the optimal (intractable) solution.

\paragraph{Limitations of existing methods.} In many data-intensive decision making applications, however, evaluating these surrogate objectives is expensive. First, let us assume that the underlying distribution over hypotheses is given. At each iteration, one needs to perform a greedy search over the tests and find the one that myopically maximizes the expected gain in the corresponding objective, whose runtime depends linearly on the size of the support of the probability distribution over the hypotheses. However, with the size of the hypothesis space
growing exponentially in the number of tests, it is often computationally prohibitive to work with the original distribution. 
Second, in practice, the underlying distribution over hypotheses is often unknown, and requires to be estimated (and learned) over time. Within an online framework for solving the troubleshooting problem, 
we assume that the virtual agent does not possess a perfect knowledge of which root-causes correspond to which symptoms. Thus, to provide better diagnosis in the long run, the virtual agent must engage customers to answer more explorative questions during each session, while not spamming the customer with excessive queries. 


\paragraph{Our contribution.}

\looseness -1 In this paper, we make contributions to both fronts. First, assuming that the prior over hypotheses is given, we propose an efficient hypothesis enumeration scheme,  which makes the class of adaptive submodular surrogates practically feasible, while still preserving strong theoretical guarantees.
In particular, through a ``divide-and-conquer'' strategy, we generate the most probable hypotheses conditioning on each hidden state with a novel and efficient \emph{priority search} procedure, and then merge them over all states to compute their marginal 
likelihood. 
In comparison with prior art, our sampling scheme utilizes the specific structure of the underlying model, and thereby offers increased efficiency and better approximation guarantees.


As our second contribution, we integrate our hypothesis enumeration strategy for optimizing VoI into an \emph{online} sequential information gathering framework, where the conditional probabilities of test outcomes given the hidden states are unknown, and can only be learned from data in an online fashion. For instance, in troubleshooting, the conditional probabilities of symptoms given a root-cause might be unknown. For this purpose, we employ a \emph{posterior sampling} approach
, where for each decision-making session (i.e., for each customer), we first sample parameters of the conditional probability distributions according to their probabilities of being ``optimal'' (in the sense that they reflect the true parameters), and then use our hypothesis enumeration algorithm to generate hypotheses for that session.

We further establish a rigorous bound on the \emph{expected regret} (defined in terms of the value of information) of our algorithm. Our online learning strategy can be interpreted as Thompson sampling across multiple sessions of interaction. Several recent empirical simulations~\cite{NIPS2011_Chapelle,GraepelCBH10,Scott:2010} and theoretical studies~\cite{AgrawalG12,NIPS2013_Bubeck,Kaufmann:2012} have demonstrated the effectiveness of Thompson sampling in different settings. However, different from our framework, 
the classical usage of Thompson sampling~\cite{Thompson1933} suggests to choose an action according to its probability of being optimal, i.e. the action which maximizes the reward in expectation; whereas in our model, the ``action'' can be interpreted as the set of tests performed in one decision-making session.

Finally, we demonstrate our online learning framework on a real-world troubleshooting platform. Our empirical results show that one can efficiently run the ``submodular surrogate''-based approaches with our dynamic hypothesis enumeration strategy, while achieving much better performance comparing with existing commonly-used heuristics (we observe a 16\% improvement on the average cost on our troubleshooting dataset). Our experiments under the online setting imply that our framework encourages efficient exploration, which, combined with the hypothesis enumeration algorithm, leads to efficient online learning of the optimal VoI.


\section{VALUE OF INFORMATION}\label{sec:voi_n_ect}



In this section we introduce basic notations, and formally state the VoI problem and existing methods for solving it.

\vspace{-2mm}\paragraph{Formulating VoI as a DRD problem.}
Let $Y\in\cY \triangleq \{y_1,\dots, y_m\}$ be a random variable that represents some \emph{hidden state}, upon which we want to make a \emph{decision}. The reward of making decision $d\in \cD$ for hidden state $y \in \cY$ is modeled by a utility function $u: \cD \times \cY \rightarrow [0,1]$. We are given a set of \emph{tests} $\cT \triangleq \{1, \dots, n\}$ of \emph{binary} outcomes; performing each test $t \in \cT$ reveals some information about $Y$, and incurs some cost which is given by a cost function $c: \cT \rightarrow \NonNegativeReals$. Let $\cH$ denote the ground set of \emph{hypotheses}; each hypothesis $h\in \cH$ corresponds to a possible realization of the outcomes of all tests in $\cT$. In other words, the \emph{outcome} of test $t$ can be modeled as a deterministic function of $h$, i.e., $g_t: \cH \rightarrow \obsDom$. Let $X_t \in \obsDom$ be the random variable corresponding to the outcome of test $t$, and $H = [X_1, \dots, X_n]$ be the random variable over $\cH$. We use $x_t$ to denote the observed outcome of test $t$. 
Crucially, we assume that $X_i$'s are \emph{conditionally independent} given the hidden state $Y$, i.e., $\Pr{Y,\Obs_1,\dots,\Obs_m} = \Pr{\Hiddenvar} \prod_{i=1}^n \Pr{X_i \given Y}$. In such \emph{offline} setting, we assume that the parameters of the above distributions are given.

Denote the set of performed tests by {$\cA$} and their outcome vector by $\bx_\cA$. We define $\expctutilof{d\given\bobs_\cA}\triangleq \expctover{\hiddenvar}{u(\hiddenvar, \decision) \given \bobs_{\cA}}$ to be the expected utility of making decision $d$ after observing $\bobs_{\cA}$. The \emph{value} of a specific set of observations $\bobs_\cA$ is then defined as:
$\valueof{\bobs_{\cA} } \triangleq \max_{\decision \in \Actionset} \expctutilof{\decision \given\bobs_\cA}$,
i.e., the maximum expected utility achievable when acting upon observations $\bobs_{\cA}$. For each decision $d\in \cD$, we define its associated \emph{decision region} as $\region_d \triangleq \left\{ h : U(d \given h) = \VoI(h) \right\}$, i.e., the set of hypotheses for which $d$ is the optimal decision.

Formally, a \emph{policy} $\policy:2^{\Testset \times \obsDom} \rightarrow \Testset$ is a partial mapping from the set of test-observation pairs to (the next) tests. The expected cost of a policy $\policy$ is
$\cost_{av}(\policy) \triangleq \expctover{h}{\sum_{i: (i,x_i) \in \played{\policy}{h}}c(i)}$, and worst-case cost is $\cost_{wc}(\policy) \triangleq \max_{h}{\sum_{i: (i,x_i) \in \played{\policy}{h}}c(i)}$, where $\played{\pi}{h}$ represents the set of tests (and their outcomes) seen by $\pi$ given that hypothesis $h$ happens. The goal of the DRD problem is to find an optimal policy $\pi^*$ with a minimal cost (expected or worst-case), such that upon termination, there exists at least one decision region that contains all hypotheses consistent with the observations acquired by the policy. Formally, we seek
\begin{align}
  \policy^* &\in \argmin_{\policy}{\cost(\policy)},\nonumber\\
            &\text{s.t.} ~\forall h ~\exists d:
              \mathcal{H}(\played{\pi}{h})\subseteq \region_d.
              \label{eq:drd}
\end{align}
where $\mathcal{H}(\played{\pi}{h})=\{h'\in\mathcal{H}: (i,x)\in \played{\pi}{h} \Rightarrow g_i(h')=x\}$ is the set of hypotheses consistent with~$\played{\pi}{h}$.

\paragraph{The \emph{Equivalence Class Edge Cutting} algorithm.}
For simplicity, we consider the special case of the DRD problem where the decision regions are disjoint\footnote{Note that our algorithmic framework can also be directly applied to the general DRD setting with overlapping decision regions. 
}. In such case, the DRD problem reduces to the \emph{equivalence class determination} problem, which can be solved near-optimally by the \emph{Equivalence Class Edge Cutting} (\ECT) algorithm \cite{golovin10near}.

\begin{figure*}[t]
  \centering
  \begin{subfigure}[b]{0.3\textwidth}
    \centering
    \includegraphics[width=\textwidth]{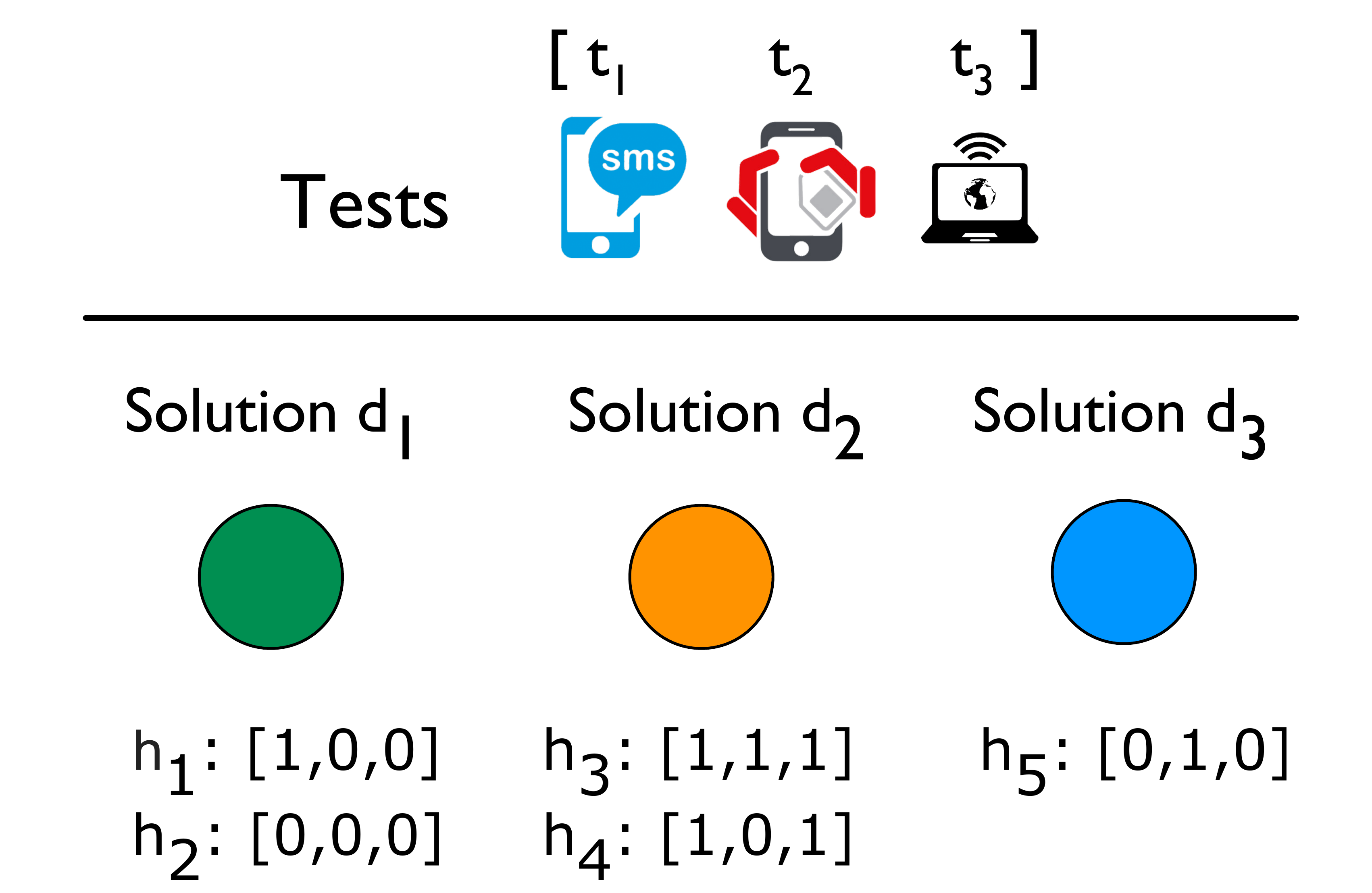}
    \caption{A troubleshooting example}
    \label{fig:ec2example}
  \end{subfigure}
  \qquad
  \begin{subfigure}[b]{0.22\textwidth}
    \centering
    \includegraphics[width=\textwidth]{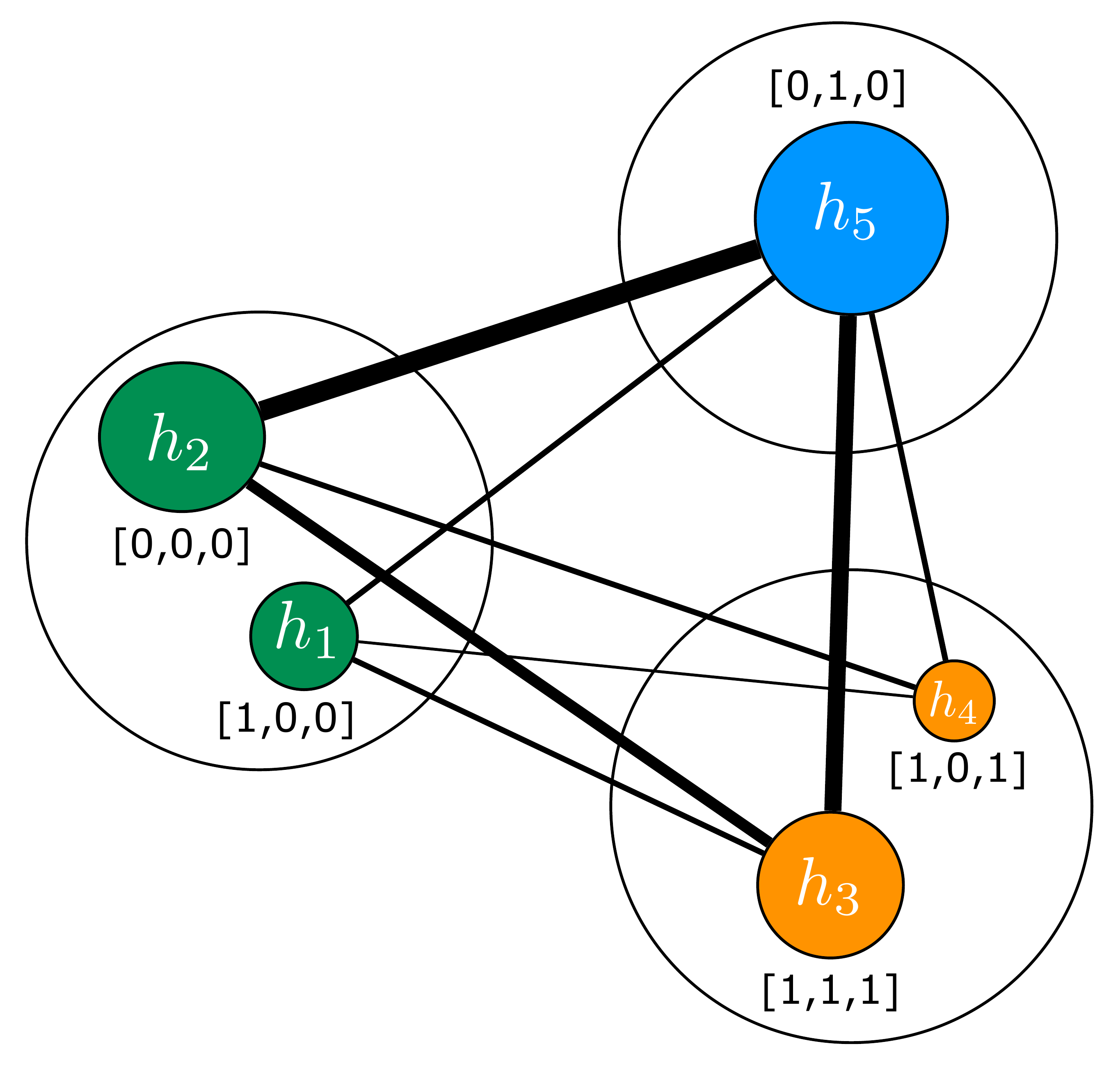}
    \caption{Initialization of \ECT}\label{fig:ecd}
  \end{subfigure}
  \qquad
  \begin{subfigure}[b]{0.22\textwidth}
    \centering
    \includegraphics[width=\textwidth]{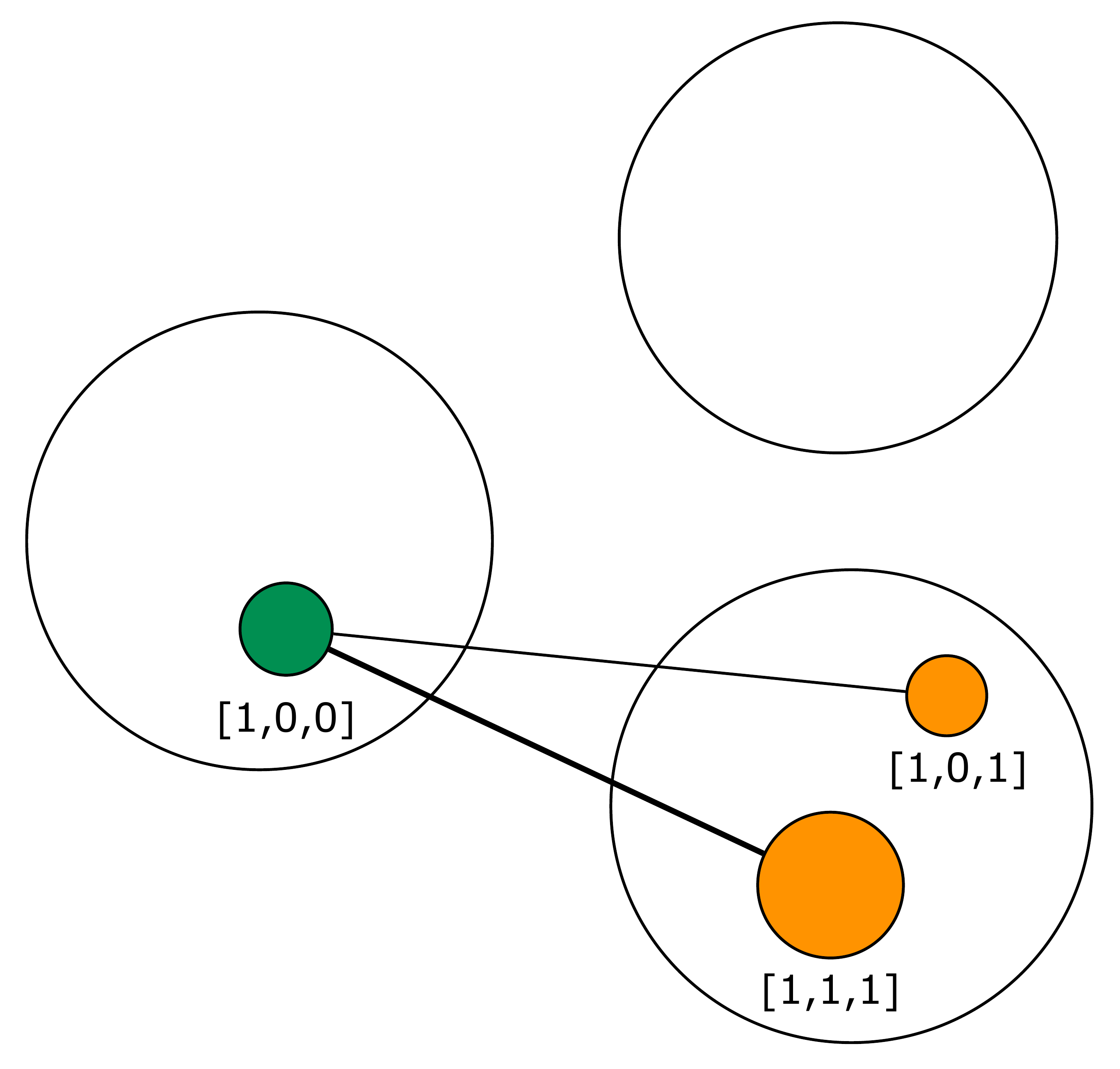}
    \caption{Observing $X_1 = 1$} \label{fig:ecd_cut}
  \end{subfigure}
  \caption{Illustration of the \ECT algorithm of \cite{golovin10near}. {Fig. (a) shows an illustrative example of the troubleshooting problem. In this example, we assume that there are three possible troubleshooting decisions we can make (i.e., solutions); each of them corresponds to one or many hypotheses (e.g., realizations of test outcomes). The goal is to identify the best solution for a given problem as quickly as possible.  In (b), we initialize \ECT, by drawing edges between all pairs of hypotheses (solid circles) that are mapped into different decisions (hollow circles). In (c), we perform test $1$ and observe $X_1 = 1$. As a result, all the edges incident to hypotheses $\hypothesis_2: [0,0,0]$ and $\hypothesis_5: [0,1,0]$ are cut/ removed.
    }} %
  \label{fig:example_ec2}
\end{figure*}

As is illustrated in \figref{fig:example_ec2}, \ECT employs an edge-cutting strategy based on a weighted graph $G = (\cH, E)$, where vertices represent hypotheses, and edges link hypotheses that we want to distinguish between. Formally, $E \triangleq \bigcup_{d\neq d'} \{\{ \hypothesis, \hypothesis' \}: \hypothesis \in \region_{d}, \hypothesis' \in \region_{d'}\}$ consists of all (unordered) pairs of root-causes corresponding to different target decisions (see \figref{fig:ecd}). We define a weight function $w: E \rightarrow \NonNegativeReals$ by $w(\{\hypothesis, \hypothesis'\}) \triangleq \Pr{\hypothesis} \cdot \Pr{\hypothesis'}$, i.e., as the product of the probabilities of its incident root-causes.
We extend the weight function on sets of edges $E' \subseteq E$, as the sum of weight of all edges $\{\hypothesis,\hypothesis'\}\in E'$, i.e., $w(E') \triangleq \sum_{\{\hypothesis, \hypothesis'\}\in E'}w(\{\hypothesis, \hypothesis'\})$. 

Performing test $t \in \Testset$ with outcome $x_t$ is said to ``\emph{cut}'' an edge, if at least one of its incident root-causes is inconsistent with $x_t$ (See \figref{fig:ecd_cut}): Formally, the set of edges cut by observing $x_t$ is $E(x_t) \triangleq \{\{ \hypothesis, \hypothesis' \}\in E: \Pr{x_t\given \hypothesis} = 0~\vee~\Pr{x_t\given \hypothesis'} = 0\}$. 
\ECT then greedily selects the test that maximizes the total expected weight of edges cut per unit cost. 
The performance guarantee of \ECT relies on the fact that the objective function (i.e., the total weight of edges cut) is \emph{adaptive submodular}, and \emph{strongly adaptive monotone} \cite{golovin2011adaptive}. In particular, let $p_{\min} \triangleq \min_{h\in\cH}\Pr{h}$ denote the minimal prior probability of any hypothesis; the expected cost of \ECT is bounded by an $\bigO{\log(1/p_{\min}) + 1}$ factor\footnote{Throughout this paper all the log’s are in base 2.} of the minimal expected cost.

\section{EFFICIENT OPTIMIZATION OF VOI VIA HYPOTHESIS ENUMERATION}\label{sec:enumeration}
%

Note that to compute the exact \ECT objective, we have to enumerate all hypotheses in $\cH$ of non-zero prior probability. The total number of hypotheses is exponential with respect to the number of tests; hence, direct application of \ECT does not scale up to several hundreds of tests or more. To facilitate efficient optimization, we must consider effective sampling schemes to explore the hypothesis space. We aim to maintain a ``confident set'' of hypotheses to efficiently approximate the \ECT objective. Concretely, we consider the following problem:




\paragraph{The optimal hypothesis enumeration problem.} Assume the prior $\Pr{Y}$ on the hidden state is known, and the prior distribution over hypotheses are fully specified by the conditional probability distribution table (CPT): $\btheta = [\theta_{ij}]_{n\times m}$, where $ \theta_{ij} \triangleq \Pr{X_i = 1 \given Y = y_j}$ for test $i \in [n]$ and hidden state $j \in [m]$. Let $\tilde{\cH}$ be the set of hypotheses sampled from the CPT.
Clearly, an ``ideal'' set $\tH$ for \ECT should be (1) rich enough to enclose promising candidates of \emph{true} underlying hypotheses, and (2) compact enough so that it excludes hypotheses that are extremely rare and ensures feasibility of the algorithm.
To this end, we define the \emph{coverage} of $\tilde{\cH}$ as its total probability mass: $Z(\tH) = \sum_{h\in{\tH}} \PrOver{ }{h}$, and the coverage of $\tilde{\cH}$ conditioning on $y$ as $Z(\tH \given y) = \sum_{h\in{\tH}} \PrOver{ }{h \given Y=y}$. We aim to attain a high coverage over $\cH$ using samples, while keeping the sample size as small as possible. Formally, to achieve $1-\eta$ coverage, we seek $$\tH^* = \argmin_{\tH: Z(\tH) \geq 1-\eta} |\tH|.$$


Existing approaches for generating hypotheses, such as Monte-Carlo sampling, often require a large sample size to reach a certain coverage of the total probability mass.
To illustrate this, let us consider a simple multinomial distribution that describes the probability distribution of four mutually exclusive hypotheses $(h_1,h_2,h_3,h_4)$, with probabilities (0.94, 0.03, 0.02, 0.01). 
A Monte-Carlo hypothesis generator simply samples hypotheses according to their probabilities (as we were rolling a dice). If we require to observe a subset of hypotheses that cover at least 98\% of the total probability mass (i.e. $h_1, h_2$ and at least one of $h_3$ or $h_4$) with a confidence level of at least 99\%, then we need at least a sample of average size 174, to cover the ``rare'' observations.  
%

\begin{figure}[t]
  \centering
  \begin{subfigure}{.45\textwidth}
    \centering
    \includegraphics[width=\textwidth]{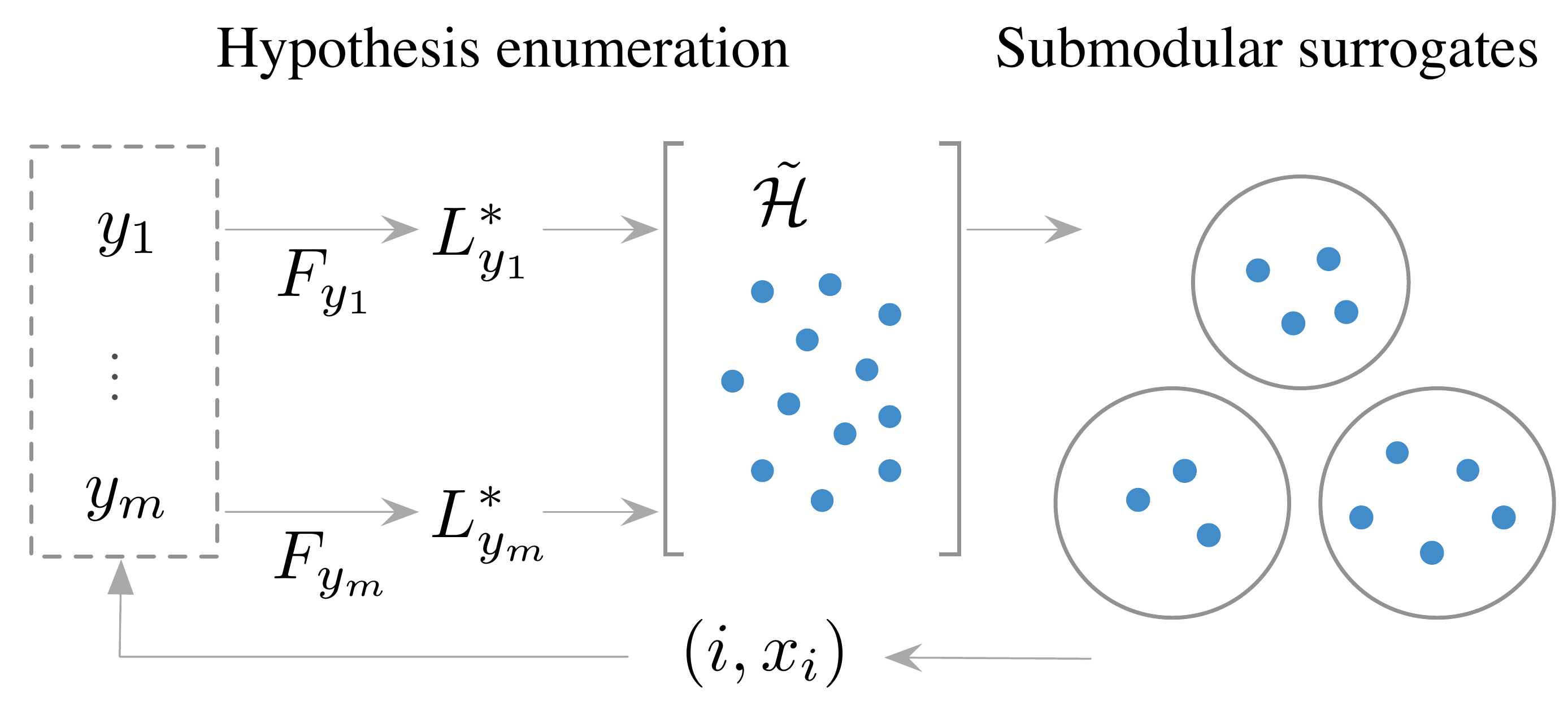}
  \end{subfigure}
  \caption{The dynamic hypothesis enumeration framework for optimizing VoI.} \label{fig:sampling}
\end{figure}

\paragraph{Dynamic hypothesis enumeration.} The problem of the Monte-Carlo approach is that it is ignorant of the structure of the VoI problem. Instead, our method aims at providing the most likely configurations -- covering up to a pre-specified fraction of the total probability mass -- in an efficient and adaptive way. In a nutshell, we adaptively maintain a pool of hypotheses that constitute a small sample with sufficient coverage.
In particular, our hypothesis enumeration scheme consists of two modules: (1) \algref{alg:local} \emph{locally} enumerates the \emph{most likely} hypotheses for each hidden state, which will cover -- by taking the union over all hidden states -- at least $(1 - \eta)$ fraction of the total probability mass of all hypotheses; and (2) \algref{alg:global}, as illustrated in \figref{fig:sampling}, provides a \emph{global} mechanism that, after observing a test outcome, adaptively filters out inconsistent hypotheses and \emph{re-generates} new hypotheses by calling \algref{alg:local}. 

\subsection{ENUMERATING HYPOTHESES}\label{sec:local}


The basic module of our hypothesis enumeration framework is a ``local'' hypothesis generator, which enumerates the most likely hypotheses for any given hidden state. It incrementally builds a Directed Acyclic Graph\footnote{Note that this DAG is used as a data structure for hypothesis enumeration, and is \emph{different} from the (undirected) weighted graph used for running the \ECT algorithm.} (DAG) of hypotheses, starting from the most likely configuration. At each step, the leaf nodes of the DAG represent the current \emph{candidate frontier}, i.e., the set of hypotheses that dominate all other candidate hypotheses in terms of likelihood. This set is used to generate the remaining hypotheses through a ``children generation'' mechanism: the next most likely hypothesis of \emph{candidate frontier} is identified, and its (at most two) children are added as new leaf nodes to the DAG.

\begin{algorithm}[t!]
  \nl {\bf Input}: {Hidden state $y$, Conditional probability table $\btheta$, coverage threshold $\eta$, (optional) frontier $F_y$}; \\
  \Begin{
    \nl Sort tests in decreasing order of $\Pr{X_i = 1\given y}$; \label{algln:reordering}\\
    \ForEach{$i \in \{1,\dots,n\}$}{
      \nl $p_i \leftarrow \log (\Pr{X_i = 1\given y})$; \label{algln:compp}\\
      \nl $q_i \leftarrow \log (\Pr{X_i = 0\given y})$; \label{algln:compq}\\
    }
    \If{$F_y$ is empty}{
      \nl $F_y \leftarrow \{ h_1=[1, 1, \dots, 1] \}$, with log-weight $\lambda_y(h_1) = \sum_i \log p_i$ ; \label{algln:fempty1}\\
      \nl $L_y^* \leftarrow \emptyset$; \label{algln:fempty2}\\
    }
    \While{$\sum_{h \in L_y^*} \exp(\lambda_y(h)) < 1 - \eta$}
    {
      \nl $h^* \leftarrow \argmax_{h\in F_y} \lambda_y(h)$; \label{algln:local_while_loop_begin}\\
      \nl $F_y \leftarrow F_y \setminus \{h^*\}$, $L_y^* \leftarrow L_y^* \cup \{h^*\}$;\\
      \nl Generate (at most) 2 children $h_{c_1}, h_{c_2}$ from $h^*$; \label{algln:local_gench}\\
      \nl $F_y \leftarrow F_y \cup \{h_{c_1}, h_{c_2}\}$; \label{algln:local_while_loop_end}\\
    }

    \nl {\bf Output}: Most likely hypotheses $L_y^*$ for $y$, log- probabilities $\lambda_y(h) = \log (\Pr{h \given y,\bx_\cA})$, and $F_y$.\\
  }
  \caption{Generate the most likely hypotheses for $y$}\label{alg:local}
\end{algorithm}

The input of \algref{alg:local} consists of the given hidden state value $y$,
the associated outcome probability vector over $n$ tests, i.e., $\Pr{x_i\given y}$ ($i = 1, \ldots, n$), and the threshold of coverage $\eta$. Optionally, it might be given a candidate frontier $F_y$, which is defined as a list of consistent hypotheses $h$ with their log-probability weights  $\lambda_y(h) = \Pr{h\given y,\bx_\cA}$ conditioned on the hidden state value $y$ and current observations $\bx_\cA$. 
$F_y$ is obtained as a by-product when calling the same module for the same $y$ at the previous iterations, and is used as a seed set of nodes to further expand the DAG. 

W.l.o.g., we can assume that tests' outcomes are defined in such a way that $\Pr{X_i=1\given y}\geq 0.5$.\footnote{Otherwise, we can redefine a test with flipped labels.} Initially (line~\ref{algln:reordering}), 
the tests are rearranged in decreasing order of 
$\Pr{X_i = 1\given y}$. Thereby, the last test will be the one with the highest uncertainty; hence flipping the sign of this test will have the minimal effect on the overall likelihood. The generator then proceeds to enumerate the most likely hypotheses corresponding to the given hidden state $y$. At line~\ref{algln:local_gench}, the two children hypotheses are generated as follows. For the first child, if the last (right-most) bit of $h^*$ is 1, we then create $h_{c_1}$ by switching the last bit to 0. For instance, the child hypothesis $h_{c_1}$ of $h^*=[0, 1, 1, 0, \mathbf{1}]$ is $[0, 1, 1, 0, \mathbf{0}]$. Its log-probability is obtained by $\lambda_y(h_{c_1}) = \lambda_y(h^*) + q_n - p_n$. For the second child, we first need to locate the right-most ``$[1,0]$'' pair in $h^*$ (if there exists any; otherwise we do nothing), and the create $h_{c_2}$ by switching ``$[1,0]$'' into ``$[0,1]$''. For instance, the child hypothesis $h_{c_2}$ of $h^*=[0, 1, \mathbf{1, 0}, 1]$ is $[0, 1, \mathbf{0, 1}, 1]$. Its associated log-probability is computed by $\lambda_y(h_{c_2}) = \lambda_y(h^*) + q_i  - p_i + p_{i+1} - q_{i+1}$, where $i$ is the bit index of the ``1'' in the right-most ``$[1,0]$'' pair.

As output, \algref{alg:local} produces a ranked list $L_y^*$ of the most likely hypotheses for a given $y$, and their log-probabilities $\lambda_y(h) = \log (\Pr{h\given y,\bx_\cA})$, such that $\sum_{h \in L_y^*} \exp(\lambda_y(h)) \ge 1 - \eta$. In addition, it also produces a residual frontier $F_y$ that will be used 
as a new ``seed'' list for the next iteration.

\subsection{ITERATIVE FILTERING AND HYPOTHESIS RE-SAMPLING }
After generating the most likely hypotheses for each hidden state, we merge them into a global set and compute their marginal likelihoods. 
We dynamically re-generate new hypotheses as more observations are made. This step is necessary in order to constantly guarantee that the sample set covers at least $1-\eta$ of the total remaining mass, after new observations become available. 

\begin{algorithm}[t!]
  \nl {\bf Input}: {Conditional probability table $\btheta$, Prior $\PrOver{}{Y}$, coverage threshold $\eta$}; \\
  \Begin{
    \nl $\tH \leftarrow \emptyset$; \\
    \While{stopping condition for \ECT not reached}{
      \ForEach{$y \in \{y_1,\dots,y_m\}$}{
        \nl Call \algref{alg:local} to generate $L^*_y$; \label{algln:gb_lc}\\
        \nl $\tH \leftarrow \tH \cup L^*$; \label{algln:gb_merge}\\
      }
      \ForEach{$h \in \tH$}{
        \nl $p(h \mid \bx_\cA) \leftarrow \sum_y \exp(\lambda_y(h))\cdot \Pr{y\given \bx_\cA}$;\\
      }
      \nl \label{alg:line:ec2} Run \ECT to determine the next test $t$; $\cA \leftarrow \cA \cup \{t\}$; \\
      \nl Observe $x_t$; $\bx_\cA \leftarrow \bx_\cA \cup \{x_t\}$; \\
      \nl Update $\Pr{y\given \bx_\cA}$; \\
      \nl $\lambda_y(h) \leftarrow \lambda_y(h) - \log \Pr{x_t\given y}$; \\
      \nl Filter out inconsistent hypotheses in $L_y^*$ and $F_y$;\\
      \nl Remove test $t$ from the list of available tests; \\
    }
    \nl {\bf Output}: (test - outcome) vectors $\bx_\cA$, decision $d$\\
  }
  \caption{Iterative Filtering and Re-sampling}\label{alg:global}
\end{algorithm}

As shown in \algref{alg:global}, the global iterative filtering and re-sampling module consists of a global loop, where after initializing all ranked lists $L_y^*$ to $\emptyset$ and $\Pr{y\given \bx_\cA=\emptyset}$ to the prior distribution over the hidden states, it iteratively performs the following sequences of operations: First, for each hidden state $y$, it calls \algref{alg:local} to generate enough hypotheses so that $L_y^*$ covers at least $(1-\eta)$ of its current mass, i.e., $Z(L_y^* \given y, \bx_\cA) \geq 1 - \eta$ (line \ref{algln:gb_lc}). $L_y^*$ might not be initially empty due to a previous call to \algref{alg:local}. In this case, the generator  produces only new additional hypotheses starting from the frontier $F_y$ until the desired coverage is achieved. This step is not necessary for the $y$'s that are inconsistent with $\bx_\cA$, i.e., for those hidden states whose posterior distribution given $\bx_\cA$ is zero.


Once we merge the hypotheses associated with each hidden state (line \ref{algln:gb_merge}), the sample set $\tH$ covers at least $(1-\eta)$ fraction of the total mass that is consistent with all the observations up to $\bx_\cA$: 
$Z(\tH \given \bx_\cA) = \sum_{h \in \tH} \Pr{h \given \bx_\cA} 
\geq \sum_y \sum_{h \in L_y^*} \Pr {h \given y,\bx_\cA} \Pr{y\given \bx_\cA} \geq \sum_y (1 - \eta) \Pr{y\given \bx_\cA} = (1 -\eta)$.
The procedure is then followed by performing \ECT 
on $\tH$ to identify the next test to be performed (see \figref{fig:sampling}).

\subsection{UPPER BOUNDS ON THE COST}\label{sec:bound_offline}
Assume that we only enumerate the hypotheses \emph{once} at the beginning of each experiment, i.e., we do not re-generate the hypotheses after observing the outcome of a test. If the underlying true hypothesis is included in the sampled set $\tH$, 
then by construction, \algref{alg:global} is guaranteed to make the optimal decision. Otherwise, with small probability it fails to output the optimal decision.
Theorem \ref{thm:hiprob_statement} states a trade-off between the size of $\tH$ and the expected cost of \algref{alg:global}.



\begin{theorem}\label{thm:hiprob_statement}
  Suppose we have generated hypotheses $\tH$ with coverage $1-\eta$. Define $\tilde{p}_{\min} = \min_{h\in \tH} \frac{\Pr{h}}{1-\eta}$. Let $\policy^g_\tH$ be the policy induced by \algref{alg:global}, $\OPT$ be the optimal policy on the original distribution of $\cH$, and $c(\cT)$ be the cost of performing all tests. 
  Then, it holds that $\cost_{av}{(\policy^g_\tH)} \leq \left( 2 \ln \left(1/\tilde{p}_{\min}\right) + 1  \right) \cost_{av}(\OPT) +  \eta \cdot c(\cT).$ Moreover, if we stop running $\pi^g_\tH$ once it cuts all edges on $\tH$, then with probability at least $1-\eta$, $\pi^g_{\tH}$ outputs the optimal decision with $\cost_{wc}{(\policy^g_\tH)}\leq \left( 2\ln (1/\tilde{p}_{\min})+1 \right)\cost_{wc}(\OPT)$.
\end{theorem}


We defer the proof to the supplemental material. 
Note that the expected cost is computed w.r.t. the original hypothesis distribution $\Pr{H \given H \in \cH}$. \thmref{thm:hiprob_statement} establishes a bound between the cost of the greedy algorithm on the samples $\tH$, and the cost of the optimal algorithm on the total population $\cH$. 
The quality of the bound depends on $\eta$, as well as the structure of the problem (which determines $\tilde{p}_{\min}$). Running the greedy policy on a larger set of samples leads to a lower failure rate, although $\tilde{p}_{\min}$ might be significantly smaller for small $\eta$.
Further, with adaptive re-sampling we constantly maintain a $1-\eta$ coverage on posterior distribution over $\cH$. With similar reasoning, 
we can show that the greedy policy with adaptively-resampled posteriors yields a lower failure rate than the greedy policy which only samples the hypotheses once at the start of the session.

\section{ONLINE LEARNING FOR OPTIMIZING VOI}

In the online setting, the exact decision making model (i.e., the conditional probability table of test outcomes given the hidden state) is unknown, and we need to learn the model as we get more feedback.
We employ an efficient \emph{posterior sampling} strategy, described as follows. Suppose that initially we have access to a prior over the model parameters, for example, in the troubleshooting application, we assume a Beta prior on the parameters $[\condprob_{ij}]_{\numtest \times \numrc}$ of the CPT, 
In the beginning of session $\ell$, we sample 
$\btheta^{(\ell)}$ from $B(\balpha^{(\ell)}, \bbeta^{(\ell)})$. In particular, for each (root-cause - symptom) pair $(y_j,x_i)$, we simply generate the parameter $\theta_{ij}^{(\ell)} = \Pr{x_i = 1\given y_j}$ from $B(\alpha_{ij}^{(\ell)}, \beta_{ij}^{(\ell)})$, where $\alpha_{ij}, \beta_{ij}$ depend on historical data. Then we run \algref{alg:global} with $\btheta^{(\ell)}$ to sequentially pick tests for session $\ell$.
When a decision making session is over, we observe $y_\ell$, together with a set of test-outcome pairs. We then update the distribution on $\btheta$ before we enter the next conversation. See \algref{alg:online} for details\footnote{For simplicity, we assume that the prior $\PrOver{}{\Hiddenvar}$ over root-causes is given. In principle, we can drop this
  assumption, and instead assume a prior over the parameters of $\PrOver{}{\Hiddenvar}$ so that we
  can sample from it, similarly as how we sample $\bcondprob \sim B(\balpha, \bbeta)$.
}.


\begin{algorithm}[h]
  \nl {\bf Input}: {$\alpha_{ij}, \beta_{ij}$ parameters of Beta distributions, prior $\PrOver{\cY}{Y}$, sessions / test scenarios $\{\text{S}_1, \dots \text{S}_k\}$}; \\
  \Begin{
    \nl Set $\alpha_{ij}^{(1)} \leftarrow \alpha_{ij}; \beta_{ij}^{(1)} \leftarrow \beta_{ij}$ for all $i,j$; \\
    \nl  \ForEach{$\ell = 1\dots k$}{
      \nl $A \leftarrow \emptyset, \bx_\cA \leftarrow \emptyset$; \\
      \nl Draw $\btheta^{(\ell)} = \{{\theta}_{ij}^{(\ell)} \sim B(\alpha_{ij}^{(\ell)}, \beta_{ij}^{(\ell)})\}$; \\
      \nl Call \algref{alg:global} to engage session $\ell$; \\
      \nl Observe $\bobs_\cA$ and hidden state $y_\ell$ with index $\vartheta$; \\
      \ForEach {$(i, x_i) \in \bx_\cA$} {
        \nl \If{$x_i=1$ }{Set $\alpha_{i \vartheta}^{\ell+1} \leftarrow \alpha_{i \vartheta}^{\ell}+1$;}
        \Else{Set $\beta_{i \vartheta}^{\ell+1} \leftarrow \beta_{i \vartheta}^{\ell}+1$;}
      }
    }
  }
  \vspace{0mm}
  \caption{Online sequential decision making}\label{alg:online}
  \vspace{0mm}
\end{algorithm}



\subsection{ONLINE REGRET BOUND}


Suppose we have drawn $\btheta(\ell)$ for session $\ell$. 
Denote the optimal policy w.r.t. this distribution by $\OPT_\ell$, and the policy induced by \algref{alg:global} by $\policy^g_{\tH_\ell}$. Further let $\tilde{p}_{\min, \ell} = \min_{h\in \tH_\ell} {\Pr{h}}/{1-\eta}$. By definition, all \emph{feasible} policies that satisfy the condition of the DRD problem (Problem~\eqref{eq:drd}) achieve the same VoI (at different costs).
Hence, \thmref{thm:hiprob_statement} implies that with probability at least $1-\eta$, running \algref{alg:global} at session $\ell$ achieves the same VoI with $\OPT_\ell$, with at most $(2\ln (1/\tilde{p}_{\min}) + 1)$ times of the optimal cost.

In principle, we want to design an adaptive policy for each session that is competitive with the \emph{hindsight-optimal} policy that knows the true distribution. We define the expected \emph{regret} of a policy $\policy^g_{\tH_\ell}$ at session $\ell$ with respect to the \emph{true} distribution as
$$     \vspace{-1mm}\tilde{\Delta}_\ell = \VoI^*(\policy^g_{\tH_\ell}) - \VoI^*(\policy^*),$$
where $\VoI^*(\pi) = \expctover{h}{\max_{\decision \in \Actionset} \expctover{\hiddenvar}{u(\hiddenvar, \decision) \given \cS(\pi, h)}}$ denotes the expected utility of policy $\policy$ w.r.t. the \emph{true} distribution to be learned, and $\policy^*$ denotes the optimal policy (w.r.t. the true distribution). Suppose we are given a fixed budget $\tau$ for performing tests in each session. Define the expected regret incurred by running policies $\{\policy_1, \dots, \policy_k\}$ 
over $k$ sessions as $\Regret(k,\tau) = \sum_{i=1}^k \tilde{\Delta}_i.$
We establish the following bound on the expected regret of running \algref{alg:online}:

\begin{theorem}\label{thm:regret:ps}
  Fix $\eta \in (0,1)$. Let $\tau = \left( 2\ln (1/\delta)+1 \right)c^{wc}_{\OPT}$, where $\delta = \min_{t} \tilde{p}_{\min, \ell}$ denotes the minimal probability of any hypothesis in the sampled distributions, and $c^{wc}_{\OPT}$ denotes the worst-case cost of the optimal algorithm over any of the $k$ sessions. Then, \algref{alg:online} achieves expected regret
  \begin{align*}
    \expctover{}{\Regret(k,\tau)} = \bigO{ \tau S \sqrt{n k \tau \log(S n k \tau)} + k\eta},
  \end{align*}
  where $S$ is the total number of possible realizations of $\tau$ tests, and $n$ is the number of tests.
\end{theorem}
\looseness -1 To prove \thmref{thm:regret:ps}, 
we view the Optimal VoI problem as optimizing a (finite horizon) Partially Observable Markov Decision Process (POMDP) over repeated episodes of a fixed horizon $\tau$. The parameter $S$ in the regret bound corresponds to the number of (reachable) belief states of the POMDP. 
Once we have established this connection, we can interpret the online learning problem as a reinforcement learning problem via posterior sampling, in a similar way to \citet{osband2013more}.
Notice that a conservative bound on $S$ in $2^{n \choose \tau}$, which is doubly-exponential in the horizon. However, in practice, the number of reachable belief states is limited by the structure of the problem (e.g., configuration of the CPTs), and hence could be far smaller. In any case, \thmref{thm:regret:ps} implies that the expected regret of \algref{alg:online} in the limit (as $k \rightarrow \infty$) is bounded by $\eta/  \tau$.

\section{EXPERIMENTAL RESULTS}
\begin{figure}[t]
  \centering
  \includegraphics[width=.45\textwidth]{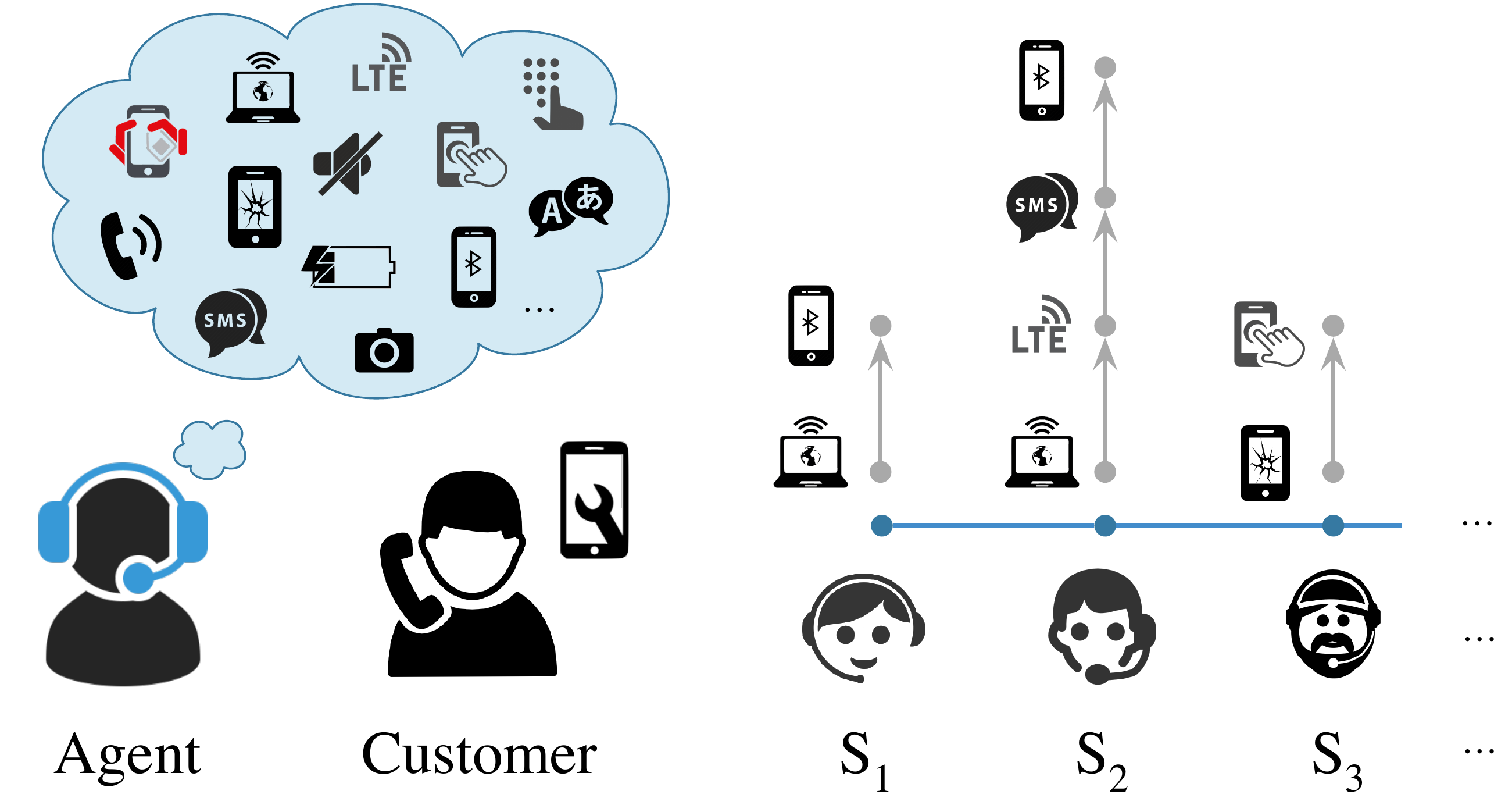}
  \caption{Online troubleshooting: customers reach a call center for diagnosis of their devices. The troubleshooting virtual agent resolves their issues by sequentially asking each customer a series of diagnosis questions.}\label{fig:online:troubleshooting}
  \vspace{-2mm}
\end{figure}


\begin{figure*}[t]
  \centering
  \begin{subfigure}{.31\textwidth}
    \centering
    \includegraphics[width=\textwidth]{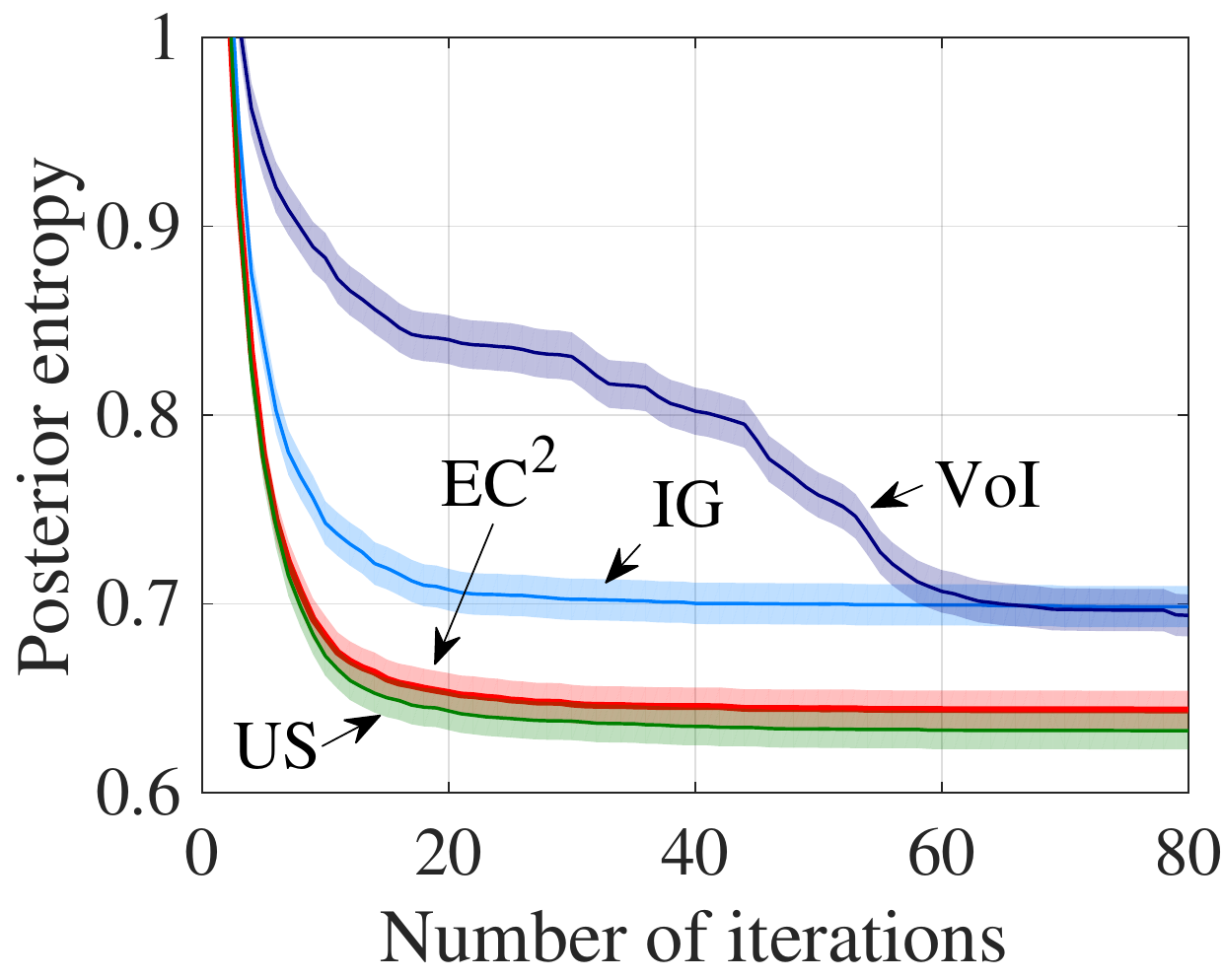} 
    \caption{Budget VS. Conditional Entropy}\label{fig:entropy}
  \end{subfigure}
  \begin{subfigure}{.305\textwidth}
    \centering
    \includegraphics[width=\textwidth]{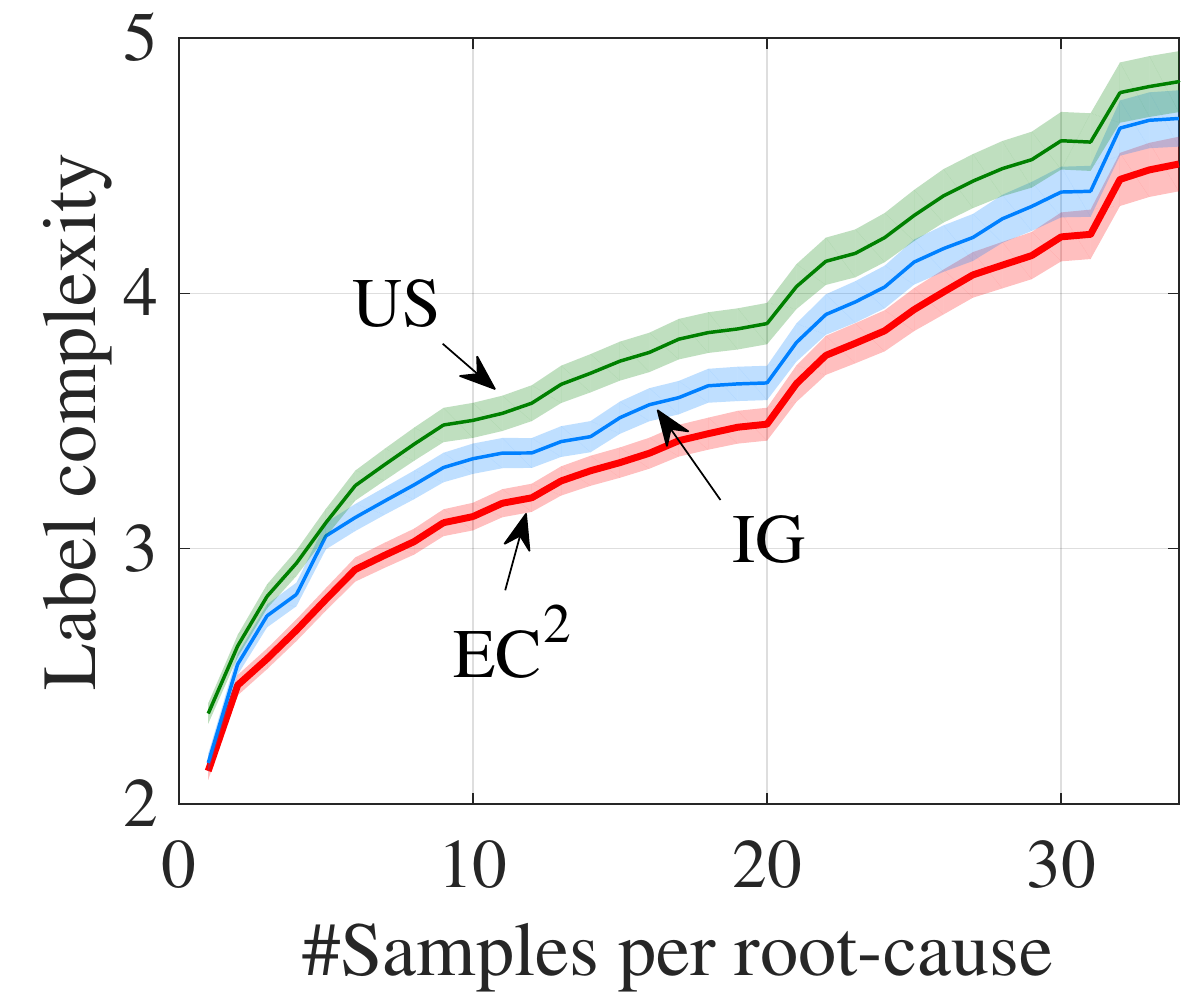}
    \caption{Cost VS. \# Samples}\label{fig:cost_vs_samples}
  \end{subfigure}
  \begin{subfigure}{.31\textwidth}
    \centering
    \includegraphics[width=\textwidth]{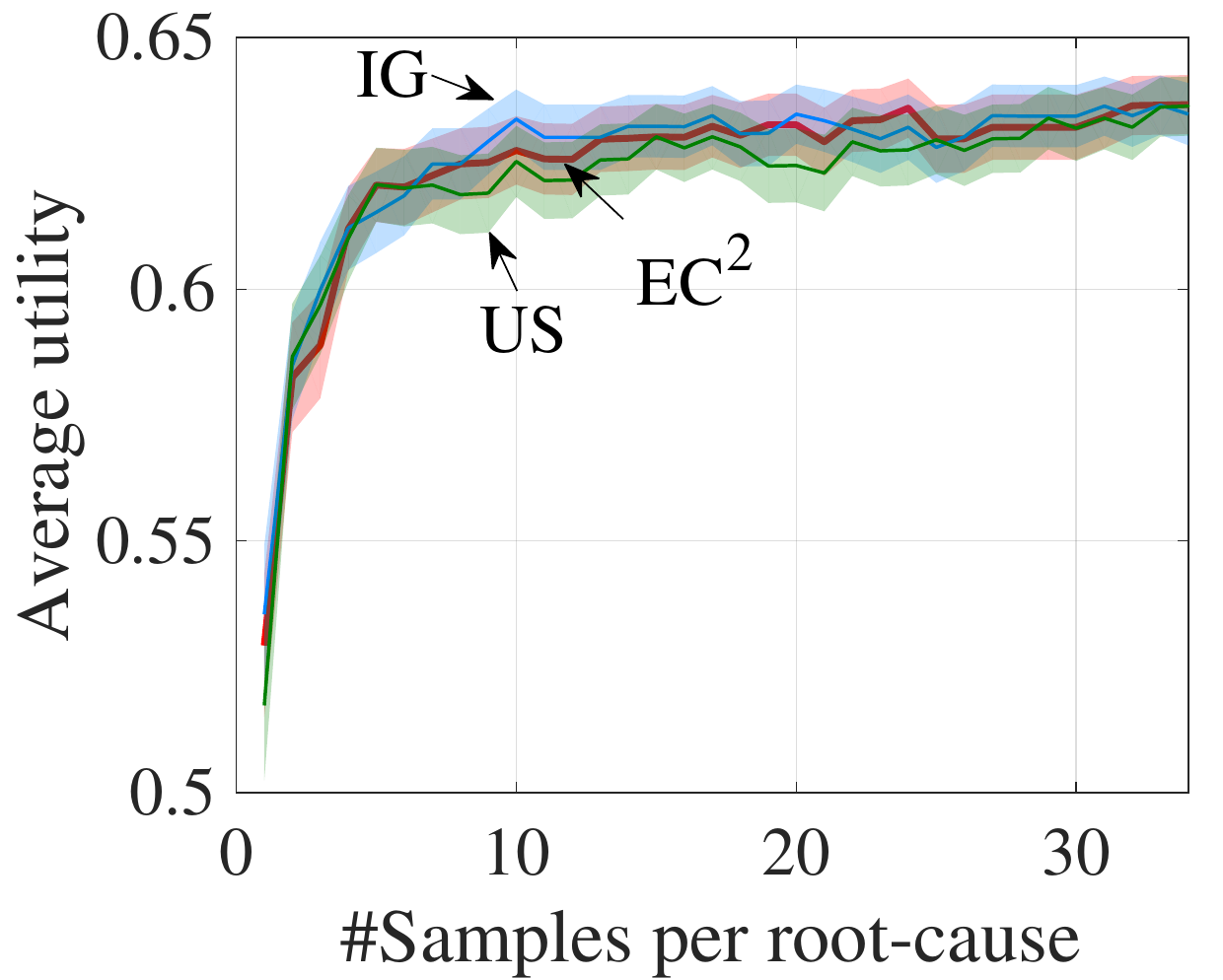}
    \caption{Utility VS. \# Samples}\label{fig:util_vs_samples}
  \end{subfigure}
  \caption{\footnotesize Experimental results by running \algref{alg:global} with different subroutines under the \emph{offline} setting.} \label{fig:movie}
\end{figure*}

\begin{figure*}[!t]
  \centering
  \begin{subfigure}{.46\textwidth}
    \centering
    \includegraphics[width=\textwidth]{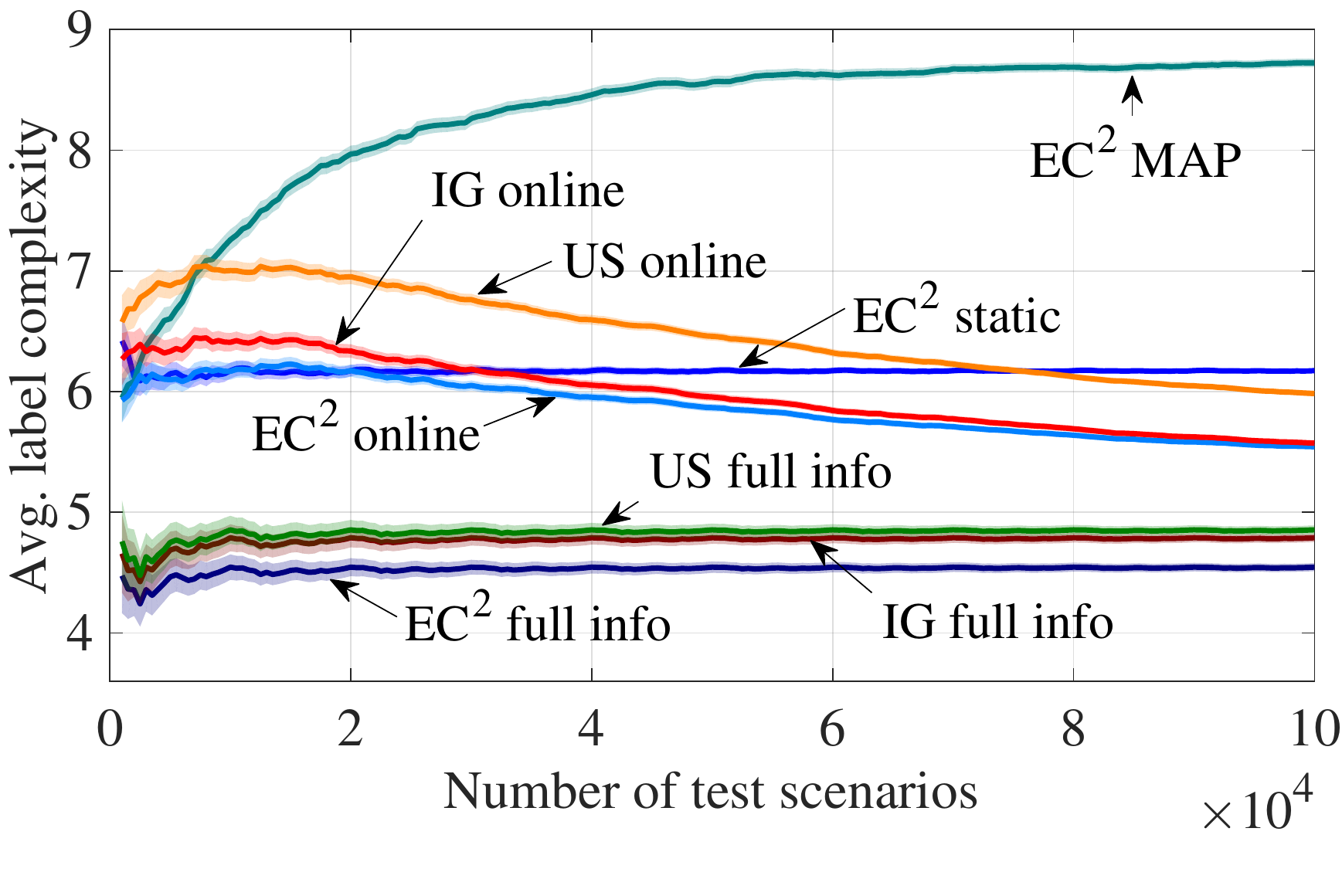}
  \end{subfigure}
  \begin{subfigure}{.46\textwidth}
    \centering
    \includegraphics[width=\textwidth]{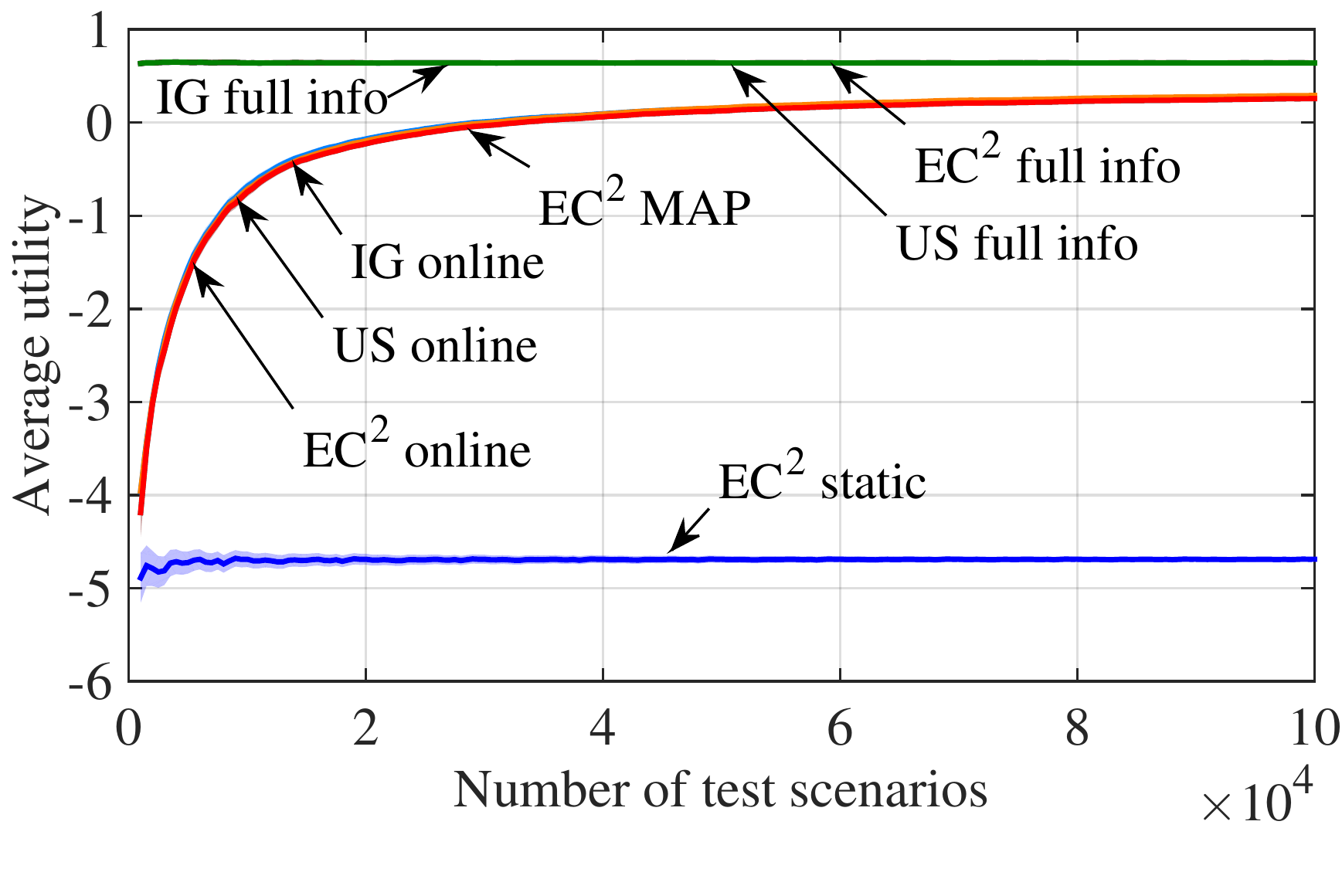}
  \end{subfigure}
  \caption{\footnotesize Experimental results by running Algorithm 3 with different subroutines under the \emph{online} setting.} \label{fig:online}
\end{figure*}

\paragraph{Data set and experimental setup.} We carry out our experiments on a \emph{real-world} troubleshooting platform. 
Our data is collected from contact center agents and knowledge workers who solve complex troubleshooting problems for mobile devices (see \figref{fig:online:troubleshooting}).
These training data involve around 1100 root-causes (the possible values of the hidden state $Y$) and 950 tests (questions on symptoms customers may encounter) with binary outcomes. From the training data, we derived a distribution over $[X_1,\dots,X_n]$ and $Y$ as $\Pr{x_1, \ldots, x_n, y} = \PrOver{\cY}{y} \prod_{i=1}^n \Pr{x_i \given y}$, where $\PrOver{\cY}{y}$ is the prior distribution over the root-causes which we assume to be uniform.

We simulated over 10,000 test scenarios (10 scenarios for each $y$), where a customer enters the system with an initial symptom 
$x_{t_0}$ (i.e. a test outcome) with probability $\Pr{x_{t_0} \given y}$. Each scenario corresponds a to a root-cause $y$ and an underlying hypothesis $h$. 
The number of decisions is the number of root-causes (which correspond to making a diagnosis), plus one extra decision of ``give-up''. Intuitively, if two root-causes result in the same symptoms, then the virtual agent cannot decide which one is the true root-cause, and therefore will forward such case to a human agent, corresponding to the ``give-up'' decision. In practice, introducing such ``give-up'' decision guarantees that there are no overlaps between decision regions. 
The utility function $u(d,y)$ corresponds to the cost of mis-prediction (either by mis-predicting a root-cause, or simply ``give-up''), which is specified by the business \emph{domain expert} as: $u(d,y) =$ (1) $0$ if $d^*$ is ``give-up''; (2) $1$ if $d^* = y$; and (3) $-19$ if $d^* \in \cY \wedge d^* \neq y$.
In this way, the ``give-up'' decision is optimal  when the posterior distribution over $Y$ given \emph{all} test outcomes has no ``peak'' value higher than 95\%. 


\paragraph{Information gathering with full knowledge of $\btheta$.}

In our experiments we set the coverage parameter $\eta=0.02$. In addition to \ECT, we also run \algref{alg:global} with several other different subroutines (by replacing \ECT at line \ref{alg:line:ec2} of \algref{alg:global}): 
myopic Value of Information ({\sc VoI}), Information Gain ({\sc IG}), and Uncertainty Sampling ({\sc US}). The myopic {\sc VoI} criterion greedily picks the test leading to the highest expected utility\footnote{Myopic VoI does \emph{not} require enumeration of hypotheses.}; the {\sc IG} (resp., {\sc US}) criterion picks tests that greedily maximizing the reduction of entropy over the decision regions (resp., hypotheses).

\figref{fig:entropy} shows the expected entropy on $Y$ while increasing the test budget. Clearly, myopic {\sc VoI} performs comparably worse than others. In \figref{fig:cost_vs_samples} and \figref{fig:util_vs_samples}, we report the average label complexity (i.e., the expected number of tests required to solve a case), and the average utility of making decisions while varying the maximal number of samples allowed for each root-cause\footnote{The \ECT, {\sc IG} and {\sc US} scores of a test can be computed efficiently in \emph{linear} time w.r.t. the number of hypotheses.}. We see that as we generate hypotheses more extensively, all algorithms require more tests in order to make a decision; on the other hand, the quality of decisions also increases with more samples. This behavior is reasonable, since having too few samples excludes a large amount of good candidates which in turn leads to poor utility.
Moreover, there is a $\sim$16\% improvement in the average query complexity when using \ECT. This shows clear advantage of using submodular surrogates for this kind of sequential problem: \ECT by construction is ``less myopic''. 

\paragraph{Online sequential information gathering}
We evaluate our online learning framework (\algref{alg:online}) over 100,000 simulated test scenarios. For the initial distribution over the CPT parameters, we set Beta priors $B(\balpha, \bbeta)$ on the CPT, where $\alpha_{ij} / \beta_{ij}$ are set to be roughly proportional to the ratio between the number of positive and negative symptoms (i.e., $(x_i=1,y_j)$ and $(x_i=0,y_j)$ pairs) in the training set. Then we inject noise into these estimates, by flipping the values of $\alpha_{ij}$ and $\beta_{ij}$ with some small probability. In our experiments, we assume a uniform prior distribution over root-causes.

\figref{fig:online} demonstrates the behavior of the three algorithms. 
In our experiments, we set a maximal budget for each session/test scenario, and keep running the policy till it identifies a decision region or exceeds the budget. At the end of session $\ell$, we report the average label complexity (i.e., the accumulated mean of the number of questions asked over $\ell$ sessions) and the average utility (i.e., the accumulated means of the utility over $\ell$ sessions). In ``\ECT MAP'', we update the CPT with their MAP estimators. We observe that its cost is much higher as it does not encourage exploration as much as the other online methods. In ``\ECT static'', we run \ECT only based on the initial sample of CPT without any updates. In the ``full info'' versions of the algorithms, the ground truth CPT is used, while in the ``online'' versions, we use \algref{alg:online} to update the CPT's. The $Y$ axis shows the accumulated means of different evaluation criteria. The two notable observations are in order: First, by integrating the two levels of sampling strategy, the average utility (a.k.a reward) of all algorithms approach the optimal utility over time.
Second, the \gls{am:ect} variants consistently outperform the alternatives in terms of query complexity, which is consistent with the results in the offline setting. 


\section{RELATED WORK} \label{related}

\paragraph{Bayesian experimental design for troubleshooting.}
There has been substantial research in diagnosis and troubleshooting using graphical models in the last two decades, in particular using Bayesian networks. One research question is how to perform efficient inference. For example, \citet{Nielsen2000} focuses on efficient belief updating in Bayesian networks, in particular in the context of troubleshooting; 
\citet{Ricks2014} consider the inference problem with more complex Bayesian network structures for multi-fault diagnosis. In contrast, we focus on an orthogonal direction, in the sense that we assume one can efficiently compute the posterior distribution over Y (i.e., the inference itself is tractable), and aim to minimize the cost of the troubleshooting policy.

\paragraph{Optimal VoI.}
Many greedy heuristics have been proposed for optimizing VoI; 
for instance, \citet{Breese1996} proposes a myopic policy for the single fault troubleshooting problem, which chooses the action/test that minimizes the expected cost of fixing the device in question. Their objective function can be viewed as a particular form of the myopic VoI. 
Unfortunately, unlike the greedy algorithms based on submodular surrogates, these algorithms can perform arbitrarily badly \cite{krause09optimal,golovin10near}. 

\paragraph{Sampling-based Bayesian inference.}
In \cite{golovin10near}, the authors suggest to use a rejection sampling approach to estimate the original distribution over hypotheses for the DRD problem. Another work, which is similar to our work in spirit, suggests to generate hypotheses in descending order of prior likelihood \cite{lerner2001inference} for approximate inference in hybrid Bayesian networks. In comparison, 
our 
sampling scheme is based on the specific structure of the underlying model, which potentially offers increased efficiency and better approximation guarantees.

\paragraph{Sequential decision making as POMDP.}
Value of information is known as a reward function in Partially Observable Markov Decision Process 
\cite{smallwoodPOMDP,kaelbling_1998_pomdp}, where each belief state represents a set of tests and their outcomes. Unfortunately, this gives us an exponentially large state space. The idea of using samples to speed up planning has been explored, 
e.g., POMCP \cite{Pineau:2006:APA:1622572.1622582} which is based on Monte Carlo tree search that samples from states and action histories, and DESPOT \cite{somani2013despot} which samples scenarios for evaluation of all policies at each iteration. 
These approaches 
rely on a finite set of policy to be evaluated. As the number of tests 
increases, 
they will require more samples. Consequently, they are limited to short planning horizons and small state and action spaces.

%
\paragraph{Online learning.}
Many different schemes have been investigated in online learning~\cite{Blum:1998} and online decision making ~\cite{Foster:1999,Kalai03efficientalgorithms}. We look into a different setting, where we integrate sequential information gathering 
into an \emph{online} learning framework wherein we learn the decision model over time. 
For this purpose, we lift our framework 
into a higher level sampling procedure, whose goal is to sample appropriately the decision model parameters. Several recent empirical simulations~\cite{NIPS2011_Chapelle,GraepelCBH10,Scott:2010} and theoretical studies~\cite{AgrawalG12,NIPS2013_Bubeck,Kaufmann:2012} have demonstrated the effectiveness of Thompson sampling in different settings. However,
different from our framework, 
the classical usage of Thompson sampling~\cite{Thompson1933} suggests to choose an action according to the probability of being optimal, i.e. the action which maximizes the reward in expectation. 

\section{CONCLUSION}

We pursue practical and efficient approaches for the challenging problem of optimal VoI: (1) For a given prior over hypothesis space, we propose a novel hypothesis enumeration strategy for efficient optimization of VoI, and show that it compensates the inefficiency of the popular ``submodular surrogate''-based greedy algorithms, while still enjoying provable theoretical guarantees; (2) when the prior is unknown, we propose an efficient, principled online learning framework with low expected regret in the long run. In addition, we demonstrate promising empirical performance of our framework on a real-world troubleshooting platform. We believe that our work is an important step towards practical and robust adaptive information acquisition systems.




\subsection*{Acknowledgments}
We would like to thank the anonymous reviewers and Christopher Dance for their helpful comments. This work was done in part while Andreas Krause was visiting the Simons Institute for the Theory of Computing. This work was supported in part by ERC StG 307036, a Microsoft Research Faculty Fellowship, an SNSF Early Postdoc.Mobility Fellowship and a Google European Doctoral Fellowship.


\clearpage
\bibliographystyle{plainnat}
\bibliography{reference}

\begin{thebibliography}{32}
\providecommand{\natexlab}[1]{#1}
\providecommand{\url}[1]{\texttt{#1}}
\expandafter\ifx\csname urlstyle\endcsname\relax
  \providecommand{\doi}[1]{doi: #1}\else
  \providecommand{\doi}{doi: \begingroup \urlstyle{rm}\Url}\fi

\bibitem[Agrawal and Goyal(2012)]{AgrawalG12}
Shipra Agrawal and Navin Goyal.
\newblock Analysis of thompson sampling for the multi-armed bandit problem.
\newblock In \emph{COLT}, 2012.

\bibitem[Blum(1998)]{Blum:1998}
Avrim Blum.
\newblock On-line algorithms in machine learning.
\newblock In \emph{Developments from a June 1996 Seminar on Online Algorithms:
  The State of the Art}, 1998.

\bibitem[Breese and Heckerman(1996)]{Breese1996}
John~S. Breese and David Heckerman.
\newblock Decision-theoretic troubleshooting: A framework for repair and
  experiment.
\newblock In \emph{UAI}, 1996.

\bibitem[Bubeck and Liu(2013)]{NIPS2013_Bubeck}
Sebastien Bubeck and Che-Yu Liu.
\newblock Prior-free and prior-dependent regret bounds for thompson sampling.
\newblock In \emph{NIPS}, 2013.

\bibitem[Chakaravarthy et~al.(2007)Chakaravarthy, Pandit, Roy, Awasthi, and
  Mohania]{chakaravarthy07decision}
V.~T. Chakaravarthy, V.~Pandit, S.~Roy, P.~Awasthi, and M.~Mohania.
\newblock Decision trees for entity identification: Approximation algorithms
  and hardness results.
\newblock In \emph{SIGMOD/PODS}, 2007.

\bibitem[Chaloner and Verdinelli(1995)]{chaloner1995bayesian}
K.~Chaloner and I.~Verdinelli.
\newblock {Bayesian experimental design: A review}.
\newblock \emph{Statistical Science}, 1995.

\bibitem[Chapelle and Li(2011)]{NIPS2011_Chapelle}
Olivier Chapelle and Lihong Li.
\newblock An empirical evaluation of thompson sampling.
\newblock In \emph{NIPS}, 2011.

\bibitem[Chen et~al.(2015)Chen, Javdani, Karbasi, Bagnell, Srinivasa, and
  Krause]{chen15submodular}
Y.~Chen, S.~Javdani, A.~Karbasi, J.~Andrew Bagnell, S.~Srinivasa, and
  A.~Krause.
\newblock Submodular surrogates for value of information.
\newblock In \emph{AAAI}, 2015.

\bibitem[Chen et~al.(2017)Chen, Hassani, and Krause]{chen16noisyal}
Y.~Chen, S.~H. Hassani, and A.~Krause.
\newblock Near-optimal bayesian active learning with correlated and noisy
  tests.
\newblock In \emph{AISTATS}, 2017.

\bibitem[Dasgupta(2004)]{dasgupta04}
S.~Dasgupta.
\newblock Analysis of a greedy active learning strategy.
\newblock In \emph{NIPS}, 2004.

\bibitem[Foster and Vohra(1999)]{Foster:1999}
Dean~P. Foster and Rakesh Vohra.
\newblock Regret in the on-line decision problem.
\newblock \emph{Games and Economic Behavior}, 29:\penalty0 7–35, 1999.

\bibitem[Golovin and Krause(2011)]{golovin2011adaptive}
D.~Golovin and A.~Krause.
\newblock Adaptive submodularity: Theory and applications in active learning
  and stochastic optimization.
\newblock \emph{JAIR}, 2011.

\bibitem[Golovin et~al.(2010)Golovin, Krause, and Ray]{golovin10near}
D.~Golovin, A.~Krause, and D.~Ray.
\newblock Near-optimal bayesian active learning with noisy observations.
\newblock In \emph{NIPS}, 2010.

\bibitem[Graepel et~al.(2010)Graepel, Candela, Borchert, and
  Herbrich]{GraepelCBH10}
T.~Graepel, J.~Q. Candela, T.~Borchert, and R.~Herbrich.
\newblock Web-scale bayesian click-through rate prediction for sponsored search
  advertising in microsoft's bing search engine.
\newblock In \emph{ICML}, 2010.

\bibitem[Howard(1966)]{howard66voi}
R.~A. Howard.
\newblock Information value theory.
\newblock In \emph{IEEE T. Syst. Sci. Cyb.}, 1966.

\bibitem[Javdani et~al.(2014)Javdani, Chen, Karbasi, Krause, Bagnell, and
  Srinivasa]{javdani14near}
S.~Javdani, Y.~Chen, A.~Karbasi, A.~Krause, D.~Bagnell, and S.~Srinivasa.
\newblock Near-optimal bayesian active learning for decision making.
\newblock In \emph{AISTATS}, 2014.

\bibitem[Kaelbling et~al.(1998)Kaelbling, Littman, and
  Cassandra]{kaelbling_1998_pomdp}
L.~P. Kaelbling, M.~L. Littman, and A.~R. Cassandra.
\newblock Planning and acting in partially observable stochastic domains.
\newblock \emph{Artificial Intelligence}, 1998.

\bibitem[Kalai and Vempala(2003)]{Kalai03efficientalgorithms}
Adam Kalai and Santosh Vempala.
\newblock Efficient algorithms for online decision problems.
\newblock In \emph{J. Comput. Syst. Sci}, pages 26--40, 2003.

\bibitem[Kaufmann et~al.(2012)Kaufmann, Korda, and Munos]{Kaufmann:2012}
Emilie Kaufmann, Nathaniel Korda, and R{\'e}mi Munos.
\newblock Thompson sampling: An asymptotically optimal finite-time analysis.
\newblock In \emph{ALT}, 2012.

\bibitem[Kosaraju et~al.(1999)Kosaraju, Przytycka, and Borgstrom]{kosaraju99}
S.~Rao Kosaraju, Teresa~M. Przytycka, and Ryan~S. Borgstrom.
\newblock On an optimal split tree problem.
\newblock In \emph{WADS}, 1999.

\bibitem[Krause and Guestrin(2009)]{krause09optimal}
A.~Krause and C.~Guestrin.
\newblock Optimal value of information in graphical models.
\newblock \emph{JAIR}, 2009.

\bibitem[Lerner and Parr(2001)]{lerner2001inference}
Uri Lerner and Ronald Parr.
\newblock Inference in hybrid networks: Theoretical limits and practical
  algorithms.
\newblock In \emph{UAI}, 2001.

\bibitem[Nielsen et~al.(2000)Nielsen, Wuillemin, Jensen, and
  Kjaerulff]{Nielsen2000}
T.~Nielsen, P.-H. Wuillemin, F.~Jensen, and U.~Kjaerulff.
\newblock Using robdds for inference in bayesian networks with troubleshooting
  as an example.
\newblock In \emph{UAI}, 2000.

\bibitem[Osband et~al.(2013)Osband, Russo, and Van~Roy]{osband2013more}
Ian Osband, Dan Russo, and Benjamin Van~Roy.
\newblock (more) efficient reinforcement learning via posterior sampling.
\newblock In \emph{NIPS}, 2013.

\bibitem[Pineau et~al.(2006)Pineau, Gordon, and
  Thrun]{Pineau:2006:APA:1622572.1622582}
Joelle Pineau, Geoffrey Gordon, and Sebastian Thrun.
\newblock Anytime point-based approximations for large pomdps.
\newblock \emph{JAIR}, 27\penalty0 (1):\penalty0 335--380, 2006.

\bibitem[Ricks and Mengshoel(2014)]{Ricks2014}
Brian Ricks and Ole~J. Mengshoel.
\newblock Diagnosis for uncertain, dynamic and hybrid domains using bayesian
  networks and arithmetic circuits.
\newblock \emph{International Journal of Approximate Reasoning}, 2014.

\bibitem[Runge et~al.(2011)Runge, Converse, and Lyons]{Runge20111214}
M.~C. Runge, S.~J. Converse, and J.~E. Lyons.
\newblock Which uncertainty? using expert elicitation and expected value of
  information to design an adaptive program.
\newblock \emph{Biological Conservation}, 2011.

\bibitem[Scott(2010)]{Scott:2010}
Steven~L. Scott.
\newblock A modern bayesian look at the multi-armed bandit.
\newblock \emph{Appl. Stoch. Model. Bus. Ind.}, 26\penalty0 (6):\penalty0
  639--658, 2010.

\bibitem[Settles(2012)]{settles.book12}
B.~Settles.
\newblock \emph{Active Learning}.
\newblock Morgan \& Claypool, 2012.

\bibitem[Smallwood and Sondik(1973)]{smallwoodPOMDP}
Richard~D. Smallwood and Edward~J. Sondik.
\newblock The optimal control of partially observable markov processes over a
  finite horizon.
\newblock \emph{Operations Research}, 21\penalty0 (5):\penalty0 1071--1088,
  1973.

\bibitem[Somani et~al.(2013)Somani, Ye, Hsu, and Lee]{somani2013despot}
Adhiraj Somani, Nan Ye, David Hsu, and Wee~Sun Lee.
\newblock Despot: Online pomdp planning with regularization.
\newblock In \emph{NIPS}, 2013.

\bibitem[Thompson(1933)]{Thompson1933}
William~R. Thompson.
\newblock On the likelihood that one unknown probability exceeds another in
  view of the evidence of two samples.
\newblock \emph{Biometrika}, 1933.

\end{thebibliography}

\appendix
\onecolumn
\section{OPTIMIZING VOI VIA SUBMODULAR SURROGATES}\label{sec:ec2supp}

We have discussed three orthogonal aspects for the optimal VoI problem, namely, (1) the \emph{sampling scheme} for hypothesis enumeration, (2) the \emph{online learning} framework, and (3) the \emph{choice of algorithms} for optimizing VoI. 
In this paper we focus on the first two aspects, and propose a general framework integrating the three components. 
Inevitably, the discussion of our algorithmic framework is grounded on existing submodular surrogate-based approaches for the VoI problem. We give more details of this class of algorithms in this section.

\subsection{SUBMODULARITY AND ITS IMPLICATIONS.} The \ECT objective function introduced in \secref{sec:voi_n_ect} is \emph{adaptive submodular}, and \emph{strongly adaptive monotone}. Formally, let $\bobs_\cA$ and $\bobs_\cB$ be two observation vectors. We call $\bobs_\cA$ a {\em subrealization} of $\bobs_\cB$, denoted as  $\bobs_\cA\preceq\bobs_\cB$, if the index set $\cA\subseteq\cB$ and $\Pr{\bobs_\cB \given \bobs_\cA} > 0$. A function $f:2^{\Testset \times \obsDom} \rightarrow \mathbb{R}$ is called {\em adaptive submodular} w.r.t.~a distribution $\mathbb{P}$, if for any $\bobs_\cA\preceq\bobs_\cB$ and any test $t$ it holds that $\Delta(t\given \bobs_\cA) \geq \Delta(t\given\bobs_\cB)$, where $\Delta(t\given \bobs_\cA) := \expctover{\obs_t}{f(\bobs_{\cA\cup\{t\}})- f(\bobs_{\cA}) \given \bobs_{\cA}}$ (i.e., ``adding information earlier helps more''). Further, function $f$ is called {\em strongly adaptively monotone} w.r.t.~$\mathbb{P}$, if for all $\cA$, $t\notin \cA$, and $\obs_t \in \obsDom$, it holds that
$ f(\bobs_{\cA}) \leq f(\bobs_{\cA\cup \{t\} })$ (i.e., ``adding information never hurts''). For adaptive optimization problems satisfying adaptive submodularity and strongly adaptive monotonicity, the policy that greedily, upon having observed $\bobs_\cA$, selects the test $t^*\in\argmax_t \Delta(t\given\bobs_\cA) / c(t)$, is guaranteed to attain near-minimal cost \citep{golovin10near}.

\subsection{GENERAL APPROACHES BASED ON SUBMODULAR SURROGATES.} It is noteworthy to mention that our results are not restricted to \ECT, and can be readily generalized to settings where regions are \emph{overlapped}. In such cases, we can use the \DIRECT algorithm \cite{chen15submodular}, and prove something similar with \thmref{thm:hiprob_statement}: in the upperbound, we get $\left(r\cdot\log(1/\tilde{p}_{\min}) + 1\right) \cost_{wc}(\OPT)$ (for the worst-case cost) and $\left(r\cdot\log(1/\tilde{p}_{\min}) + 1\right) \cost_{av}(\OPT) + \eta c(\cT)$ (for the average-case / expected cost), where $r$ measures the amount of ``overlap''. The analysis follows closely from the proof of \thmref{thm:hiprob_statement} in \secref{sec:proofs}. More generally, \thmref{thm:hiprob_statement} (with modified multiplicative constant) also applies to greedy algorithms whose objective function (1) is adaptive submodular, and (2) rely on a finite set of hypotheses. Other examples satisfying these conditions include \GBS \cite{golovin2011adaptive} and \HEC \cite{javdani14near}. 

Furthermore, since our framework is orthogonal to the choice of the submodular surrogate-based algorithms, we can also easily extend our analysis to handle the (more) practical setting where test outcomes are \emph{noisy}. In such settings, one can no longer ``cut-away'' edges as suggested by the \ECT algorithm, since with noisy observations one cannot ``eliminate'' any of the hypotheses (i.e., setting their probability mass to $0$). In practice, after observing the outcome of a test, we can perform Bayesian updates on the posterior over $H$ instead of eliminating those hypotheses that are ``inconsistent'' with the observation. Analogous to the analysis of \thmref{thm:hiprob_statement}, we can establish a bound on the \emph{worst-case} cost of such greedy policy, based on the recent results of \cite{chen16noisyal}. Since the theoretical question of handling noisy tests is beyond the scope of this paper, we omit the proof details for this setting.

\section{IMPLEMENTATION DETAILS OF \ALGREF{alg:local}}
In the main paper, we have stated the pseudo code of our dynamic hypothesis enumeration algorithms. Due to space limit we only provide a concise description of the main idea behind the framework. In this section, we elaborate \algref{alg:local} by providing more intuitions and implementation details, as well as additional clarifications for better understanding of the algorithm.

\paragraph{The DAG.} 
We use a DAG represents the hypotheses enumeration process, and is \emph{not} a data structure which we actually adopted in \algref{alg:local}. Rather, algorithmically we are maintaining a ``candidate frontier'' $F_y$ for each hidden state $y$, which corresponds to the set of ``leaf'' hypotheses (i.e., nodes of the DAG which have no outgoing edges), as a seed set to generate more hypotheses. \algref{alg:local} enumerates hypotheses in decreasing order of probabilities $\Pr{h\given y}$. The directed edges in the DAG indicate the relations between the conditional probabilities: if there is a directed edge from node $h_1$ to $h_2$ in the DAG, it indicates that $\Pr{h_1 \given y} \geq \Pr{h_2 \given y}$. Other than this, we do not use the edge for any other purposes.

\paragraph{Implementation details.} 
In practice, the underlying distributions are often highly concentrated, such that a few number of hypotheses cover a significant part of the total mass. On the other hand, there are many configurations with very small (but non-null) probabilities. \algref{alg:local} exploits such structural assumption, and generates the most likely hypotheses in the following four steps:
\begin{itemize}
\item \emph{Step} 1 (line~\ref{algln:reordering}): \\ Test definitions are possibly switched, in a way that $\Pr{X_i=1\mid y} \geq 0.5 \quad \forall i$ (i.e., when $\Pr{X_i=1\given y} < 0.5$,  we consider the complementary event $\bar{X}_i$ as the new test outcome so that $\Pr{\bar{X}_i=1\given y} = 1 - \Pr{{X}_i=1\given y} \geq 0.5$); test indices are re-ranked in decreasing order of  $\Pr{{X}_i=1\given y} $;
\item \emph{Step} 2 (line~\ref{algln:compp}, \ref{algln:compq}): \\ For $i=1, \ldots, n$, compute $p_i \triangleq \log (\Pr{X_i = 1\given y})$, and $q_i \triangleq \log (\Pr{X_i = 0\given y})$;
\item \emph{Step} 3 (line~\ref{algln:fempty1}, \ref{algln:fempty2}): \\ If $F_y$ is empty, initialize $F_y$ with the configuration $h_1=[1 \dots 1]$ with log-weight $\lambda_y(h_1) = \sum_i p_i$ ; set $L_y^* = \emptyset$.
\item \emph{Step} 4 (line~\ref{algln:local_while_loop_begin}-\ref{algln:local_while_loop_end}): while $\sum_{h \in L_y^*} \exp(\lambda_y(h)) < (1 - \eta)$
  \begin{itemize}
  \item \emph{Step} 4a: Choose the element $h^*$ from $F_y$ such that $\lambda_y(h^*)$ is maximum;
  \item \emph{Step} 4b: Remove $h^*$ from $F_y$ and push it into $L_y^*$;
  \item \emph{Step} 4c (line~\ref{algln:local_gench}): Generate (at most) 2 children from $h^*$ and add them to $F_y$ if they were not already present in $F_y$.
  \end{itemize}
\end{itemize}
In the main paper, we have given detailed description of how to generate the two children configurations ($h_{c_1}$ and $h_{c_2}$) in Step 4c. We provide some additional insight to facilitate better understanding of the procedure:
\begin{itemize}
\item \emph{Child} 1: Once we have re-ranked the tests in decreasing order of $\Pr{X=1 \given y}$ in Step 1, the last test in the ordered list will have the smallest probability (conditioning on $y$) of being realized to its more likely outcome, and hence is the most uncertain one. If follows that if we flip the outcome of such test, we will generate a new hypothesis $h$ with the highest $\Pr{h \given y}$ among the unseen hypotheses.
The first child is generated exactly in this way: if the last (right-most) bit of $h^*$ is 1, we then create $h_{c_1}$ by switching the last bit to 0. For instance, the child hypothesis $h_{c_1}$ of $h^*=[0, 1, 1, 0, 1]$ is $[0, 1, 1, 0, 0]$. Its log-probability is obtained by $\lambda_y(h_{c_1}) = \lambda_y(h^*) + q_n - p_n$. 

\item \emph{Child} 2: Besides flipping the last bit, the next most-likely hypothesis can also be the one with two bit edits of an existing hypothesis: Find the right-most ``$[1,0]$'' pair in $h^*$ (if there exists any; otherwise we do nothing), and the create $h_{c_2}$ by switching ``$[1,0]$'' into ``$[0,1]$''. For instance, the child hypothesis $h_{c_2}$ of $h^*=[0, 1, 1, 0, 1]$ is $[0, 1, 0, 1, 1]$. Its associated log-probability is computed by $\lambda_y(h_{c_2}) = \lambda_y(h^*) + q_i  - p_i + p_{i+1} - q_{i+1}$, where $i$ is the bit index of the ``1'' in the right-most ``$[1,0]$'' pair.
\end{itemize}


\section{PROOFS OF THE MAIN THEOREMS}\label{sec:proofs}

\subsection{PROOF OF \THMREF{thm:hiprob_statement}}\label{sec:proof_thm_hiprob}

In this section, we provide proofs for the upper bounds on the cost of \algref{alg:global}. In the analysis, we assume that we only sample the hypotheses \emph{once} in the beginning of each experiment (i.e., we don't resample after each iteration). 
\begin{proof}
  The main idea of the proof is illustrated in \figref{fig:sampled_set}.
  \begin{figure}[h!]
    \centering
    \includegraphics[width=0.35\textwidth]{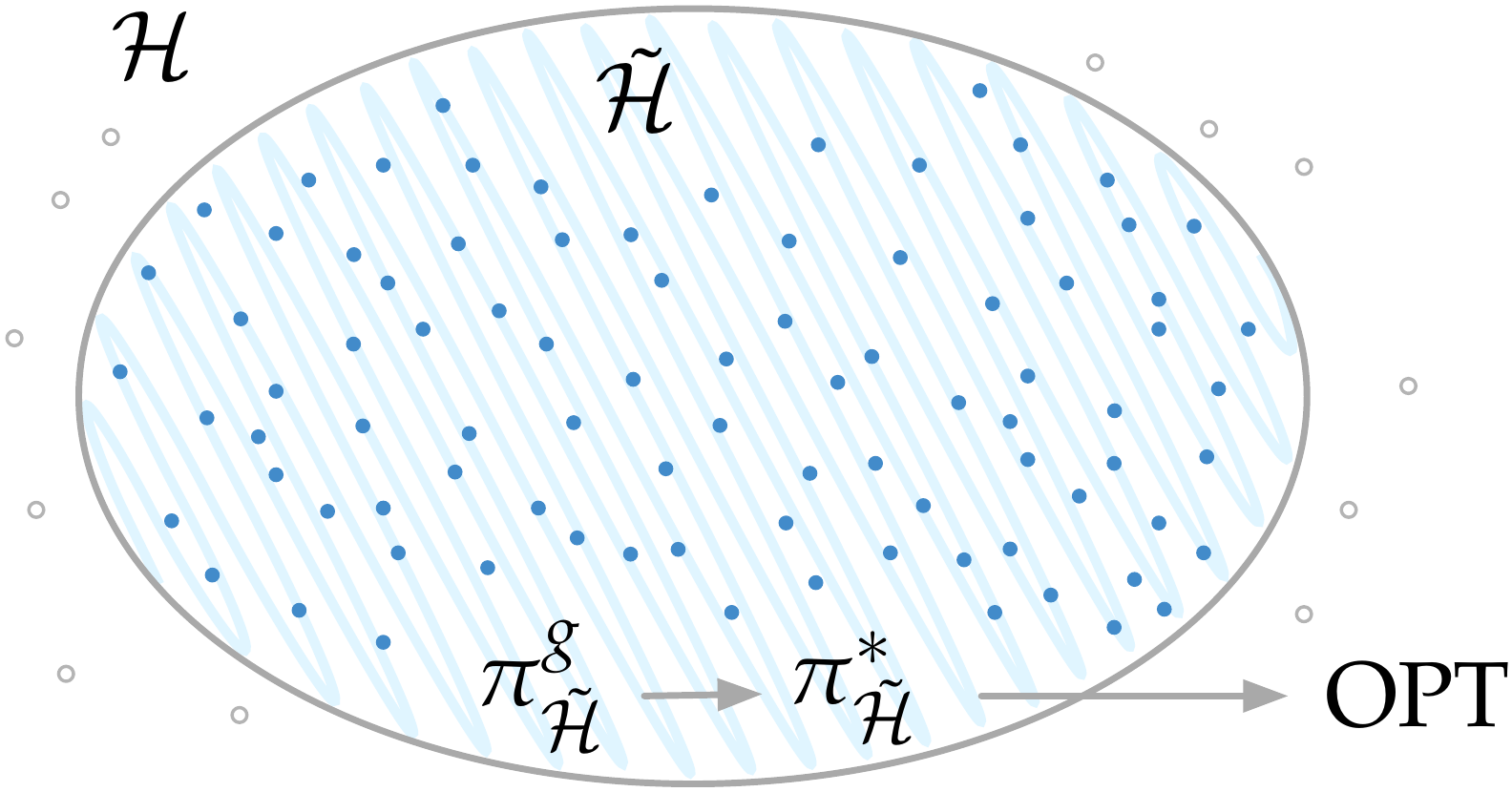}
    \caption{{Depicting the main idea behind the proof.   We introduce $\policy^*_{\tH}$ (the
        optimal policy on the sampled distribution) as an auxiliary policy to connect
        $\policy^g_{\tH}$ with \OPT. If the realized hypothesis $h^*\in\tilde{\mathcal{H}}$, then
        $\pi^g_{\tilde{\mathcal{H}}}$ efficiently identifies the decision. Otherwise, (with
        probability at most $\eta$) $\pi^g_{\tilde{\mathcal{H}}}$ randomly chooses tests, and the
        cost can be at most $c(\cT)$.
      }} %
    \label{fig:sampled_set}
  \end{figure}

  \paragraph{Bound on the expected cost}
  We first prove the upper bound on the expected cost of the algorithm.
  We use $p$ to denote the true distribution over the hypotheses $h\in \cH$, and $\tilde{p}$ be the
  sampled distribution. That is, $p(h) = \Pr{h}$, and
  \begin{align}
    \tilde{p}(h) =
    \begin{cases}
      \Pr{h} / (1 - \eta), & \text{for } h \in \tH; \\
      0, & \text{otherwise.}
    \end{cases}\label{eq:def_p_tilde}
  \end{align}
  For any policy $\policy$, let
  $\cost_{\tilde{p}} (\policy) \triangleq \expctover{h\sim \tilde{p}(h)}{c(\cS(\pi,h))}$ denote the
  expected cost of $\pi$ w.r.t.
  $\tilde{p}$. 
  Then, the expected cost of $\pi$ w.r.t. the true distribution $p$ satisfies
  \begin{align}
    \cost& (\policy) = \sum_{h\in \cH} p(h) c(\cS(\pi,h))
           \nonumber \\
         &= \sum_{h\in \tH} p(h) c(\cS(\pi,h)) + \sum_{h\in \cH\setminus\tH} p(h) c(\cS(\pi,h))
           \nonumber\\
         &\stackrel{ \text{Eq.~\eqref{eq:def_p_tilde}}}{=} (1-\eta) \sum_{h\in \tH} \tilde{p}(h) c(\cS(\pi,h)) + \sum_{h\in \cH\setminus\tH} p(h)c(\cS(\pi,h))
           \nonumber \\
         &\stackrel{}{=} (1-\eta) \cost_{\tilde{p}}(\pi) + \sum_{h\in \cH\setminus\tH} p(h)\underbrace{c(\cS(\pi,h))}_{\leq c(\cT)}
           \label{eq:upperbound_empirical_expectedcost} \\
         &\stackrel{}{\leq} (1-\eta) \cost_{\tilde{p}}(\pi) + \eta \cdot c(\cT). \label{eq:lowerbound_empirical_expectedcost}
  \end{align}
  The second term on the RHS of Eq.~\eqref{eq:upperbound_empirical_expectedcost} is non-negative,
  which gives
  \begin{align}
    {(1-\eta)} \cost_{\tilde{p}}(\pi) &\stackrel{}{=} {\cost(\policy) - \sum_{h\in \cH\setminus\tH} p(h)c(\cS(\pi,h))} \nonumber \\& \leq \cost(\policy)  \label{eq:upperbound_expcost}
  \end{align}

  Let $\pi^*_{\tilde{p}}$ be the optimal policy w.r.t. the sampled distribution $\tilde{p}$. By
  Theorem 3 of \cite{golovin10near} we get
  \begin{align}
    \cost_{\tilde{p}}\left(\pi^g_{\tH}\right) &\leq  \left( 2 \ln \left(1/\tilde{p}_{\min}\right) + 1  \right) \cost_{\tilde{p}}\left( \pi^*_{\tH} \right). \label{eq:adasubm_cost}
  \end{align}
  Therefore,
  \begin{align*}
    \cost(\pi^g_{\tH}) &\stackrel{\text{Eq.}~\eqref{eq:lowerbound_empirical_expectedcost}}{\leq}  (1-\eta) \cost_{\tilde{p}}(\pi^g_{\tH}) + \eta \cdot c(\cT) \\
                       &\stackrel{\text{Eq.}~\eqref{eq:adasubm_cost}}{\leq}  (1-\eta) \left( 2 \ln \left(1/\tilde{p}_{\min}\right) + 1  \right) \cost_{\tilde{p}}\left( \pi^*_{\tH} \right) \\ & \qquad~~ + \eta \cdot c(\cT).
  \end{align*}
  By definition we know $\cost_{\tilde{p}}\left( \pi^*_{\tH} \right) \leq \cost_{\tilde{p}}(\OPT)$. Hence
  \begin{align*}
    \cost(\pi^g_{\tH}) &\stackrel{}{\leq} (1-\eta) \left( 2 \ln \left(1/\tilde{p}_{\min}\right) + 1  \right) \cost_{\tilde{p}}(\OPT) + \eta \cdot c(\cT) \\
                       &\stackrel{\text{Eq.}~\eqref{eq:upperbound_expcost}}{\leq} \left( 2 \ln \left(1/\tilde{p}_{\min}\right) + 1  \right) \cost(\OPT) + \eta \cdot c(\cT),
  \end{align*}
  which completes the first part of the proof.

  \paragraph{Bound on the worst-case cost.} Next, we provide the proof for bound on the worst-case cost.  
  Analogous to the previous analysis, we consider two possible scenarios: (i) the realized hypotheses (i.e., the full realization vector) $h^* \in \tH$; and (ii) $h^* \notin \tH$. 



  For any policy $\policy$, the worst-case cost of $\pi$  satisfies
  \begin{align}
    \cost_{wc}(\policy) &= \max_{h\in \cH} c(\cS(\pi,h)) \nonumber\\
                        &= \max\{\max_{h\in \tH} c(\cS(\pi,h)), \max_{h\in \cH\setminus\tH} c(\cS(\pi,h))\}. \nonumber
  \end{align}
  Since policy $\pi^g_{\tilde{\mathcal{H}}}$ terminates if there is no edge left on $\tH$, then $\max_{h\in \cH\setminus\tH} c(\cS(\pi,h)) \leq \max_{h\in \tH} c(\cS(\pi,h)$. Therefore,
  \begin{align}
    \cost_{wc}(\pi^g_{\tilde{\mathcal{H}}})
    &= \max_{h\in \tH} c\left( \cS\left( \pi^g_{\tilde{\mathcal{H}}},h \right) \right) \nonumber \\
    &\stackrel{\text{(a)}}{\leq} \left( 2 \ln \left(1/\tilde{p}_{\min}\right) + 1  \right) \max_{h\in \tH}c\left( \cS\left( \pi^*_{\tilde{\mathcal{H}}},h \right) \right) \nonumber \\
    &\leq \left( 2 \ln \left(1/\tilde{p}_{\min}\right) + 1  \right) \max_{h\in H} c\left( \cS\left( \OPT,h \right) \right). \nonumber
  \end{align}
  Step (a) in the above equation follows from Theorem A.12 of \cite{golovin2011adaptive}.

  Therefore, when $\pi^g_{\tilde{\mathcal{H}}}$ terminates, with probability at least $1 - \eta$, it succeeds to output the correct decision with cost $\left( 2 \ln \left(1/\tilde{p}_{\min}\right) + 1  \right) \cost_{wc}(\OPT)$.
\end{proof}

\subsection{PROOF OF \THMREF{thm:regret:ps}}
In this section, we prove the bound on the expected regret of our online learning algorithm.
\begin{proof}[Proof of \thmref{thm:regret:ps}]
  One way to model the non-myopic value of information problem is to view it as a (finite horizon) Partially Observable Markov Decision Process (POMDP), where each  (belief-) state represents the selected tests and observed outcome of each test. Formally, the POMDP can be written as
  \begin{align}
    M \triangleq \left( \cB, \cT, R^M, P^M, \tau, \rho \right). \label{eq:pomdp}
  \end{align}

  Here, $\cB$ is the set of belief states, $\cT$ is the set of actions (i.e., tests), $R_t^M(b)$ is the (expected) reward associated with action $t$ while in belief state $b$, $P_t^M(b' \mid b)$ denotes the probabiliy of transitioning to state $b'$ if action $t$ is selectedwhile in state $b$, $\tau$ is the time horizon for each session, and $\rho$ is the initial belief state distribution.

  In our problem, the \emph{transition probabilities} $P^M$ can be fully specified by the conditional probabilities of the test outcomes given the hidden state $\Pr{x_t \given y}$; the \emph{prior} distribution $\rho$ on belief states can be specified by the prior distribution on the hypotheses $\Pr{y}$, and $\Pr{x_t \given y}$. The \emph{reward} $R^M$ for running a policy $\policy$ on $M$ is the utility achieved upon termination of the policy. More specifically, we can interpret the reward function $R^M$ as follows: we get reward $0$ as the policy keeps selecting new tests, but get (expected) reward $\VoI(\cS(\policy, h)) \triangleq \max_{\decision \in \Actionset} \expctover{\hypothesis}{u(\hypothesis, \decision) \given \cS(\policy, h)}$ if the policy terminates upon observing $\cS(\policy, h)$ and suggests a decision.
  The reward function measures the expected (total) utility one can get by making a decision after running policy $\pi$.


  We now consider running \algref{alg:online} over $k$ sessions of fixed duration $\tau$. Following the previous discussion, the problem is equivalent to learning to optimize a random finite horizon POMDP of length $\tau$ in $k$ repeated episodes of interaction. To establish the regret bound of \thmref{thm:regret:ps}, we need the following result:

  \begin{theorem}[Theorem 1 of \cite{osband2013more}]\label{thm:osband13}
    Consider the problem of learning to optimize a random finite horizon (PO)MDP $M = \left( \cB, \cT, R^M, P^M, \tau, \rho \right)$ in $k$ repeated episodes, and consider running the following algorithm: at the start of each episode it updates the prior distribution over the MDP and takes one sample from the posterior, and then follows the policy that is optimal for this sampled MDP. For any prior distribution on the MDPs, it holds that
    $$\expctover{}{\text{Regret}(k, \tau)} = \bigO{\tau |\cB| \sqrt{k\tau|\cT|\log(k\tau|\cB||\cT|)}}.$$
  \end{theorem}
  \thmref{thm:osband13} implies that the posterior sampling strategy as employed in \algref{alg:online} allows efficient learning of the MDP, given that one can find the \emph{optimal} policy for the sampled MDP at each episode. However, since finding the optimal policy is NP-hard, in practice we can only \emph{approximate} the optimal policy. In \algref{alg:online}, we consider running the greedy policy (i.e., \algref{alg:global}) in each episode to solve the sampled MDP:
  \begin{corollary}\label{cor:greedyregret}
    Let $M$ be a sampled MDP, and $c^{wc}_{\OPT}$ be the worst-case cost of the optimal algorithm on $M$. Consider running \algref{alg:global} for $\tau = \left( 2\ln (1/\delta)+1 \right)c^{wc}_{\OPT}$ steps. Then, with probability at least $1-\eta$, it achieves the optimal VoI on $M$.
  \end{corollary}
  \begin{proof}[Proof of \corref{cor:greedyregret}]
    By \thmref{thm:hiprob_statement}, we know that the greedy policy finds the target decision region with probability at least $1-\eta$. Furthermore, by definition we know that each decision region $\region_d=\{h: U(d\mid h) = \valueof{h}\}$ represents an optimal action for any of its enclosed hypotheses. In other words, a policy that successfully outputs a decision region achieves the optimal VoI.
  \end{proof}

  Denote the optimal policy on the sampled MDP in episode $i$ as $\OPT_i$. From \corref{cor:greedyregret}, we know that \algref{alg:global} achieves optimal utility with probability at least $1-\eta$ . Hence, the expected ``regret'' of \algref{alg:global} over $\OPT_i$ is
  \begin{align}
    \text{Reg}(\text{\algref{alg:global}}) \stackrel{\text{(a)}}{\leq} (1-\eta) \cdot 0 + \eta \cdot 1 = \eta. \label{eq:alg2reg}
  \end{align}
  Here, Step (a) is due to the fact that the utility is normalized so that $U\in[0,1]$. Note that $\text{Reg}(\text{\algref{alg:global}})$ in Equation \eqref{eq:alg2reg} refers to the difference between the value of \algref{alg:global} and the value of the optimal policy on the sampled MDP (not the optimal policy for the true MDP). In other words, the price of not following the optimal policy is at most $\eta$.

  By \thmref{thm:osband13}, we know that following $\OPT_i$ for episode $i$ achieves expected regret $\bigO{\tau |\cB| \sqrt{k\tau|\cT|\log(k\tau|\cB||\cT|)}}$. Further, we know that the price of approximating the optimal policy at episode $i$ is at most $\eta$. Combining these two results we get
  \begin{align*}
    \expctover{}{\text{Regret}(k, \tau)}
    &= \bigO{\tau |\cB| \sqrt{k\tau|\cT|\log(k\tau|\cB||\cT|)}} + \sum_{i=1}^{k} \eta \\
    &= \bigO{\tau |\cB| \sqrt{k\tau|\cT|\log(k\tau|\cB||\cT|)} + \eta k},
  \end{align*}
  where $|\cB| = S$ represents the number of the belief states, $|\cT| = n$ represents the number of tests. Hence it completes the proof.
\end{proof}

\section{ADDITIONAL RESULTS}
In \secref{sec:bound_offline} of the main paper (i.e., Upper Bounds on the Cost), we have provided upper bounds on the expected/worst-case cost of the greedy policy w.r.t. \emph{non-adaptively} sampled prior. In this section, we provide preliminary results for the case with adaptive re-sampling, where we constantly maintain a $1-\eta$ coverage on posterior distribution over $\cH$.
\subsection{LOWER BOUND ON THE EXPECTED \ECT UTILITY}


\begin{theorem}\label{thm:maximization_bound}
  Let $k,\ell$ be positive integers\footnote{If we assume unit cost for all tests, then $k,\ell$ are the number of tests selected. Otherwise, with non-uniform test costs, $k,\ell$ are the budget on the cost of selected items.}, $f$ be the \ECT objective function, $\policy^g_{\tH,[\ell]}$ be the greedy policy with budget $\ell$ on $\tH$, and $\policy^*_{\cH,[k]}$ be the optimal policy that achieves the maximal expected utility under budget $k$ on $\cH$, Then,
  \begin{align*}
    f_{avg}\left( \policy^g_{\tH,[\ell]} \right) \geq \left( 1- e^{-\ell/k} \right)f_{avg}\left( \policy^*_{\cH, [k]} \right) - k\epsilon,
  \end{align*}
  where $\epsilon = 2\eta \left( 1-\left(\frac{1}{k}\right)^\ell \right)$, and $f_{\text{avg}}(\pi) \triangleq \expctover{h}{f(\cS(\pi, h))}$ denotes the expected utility of running policy $\pi$ w.r.t. the original distribution.
\end{theorem}
Note that the above result applies to the \ECT algorithm with \emph{adaptive-resampled} posteriors at each iteration. The additive term $k\epsilon$ on the RHS is due to the incompleteness of the samples provided by the sampling algorithm. The main intuition behind the proof is that, due to the effect of resampling, the expected one-step gain of the greedy policy $\policy^g_{\tH,[\ell]}$ on the sampled distribution suffers a small loss at each iteration, comparing to the greedy algorithm on the true distribution. The loss will be accumulated after $\ell$ rounds, leading to a cumulative loss of up to $k\epsilon$ in the lower bound.

We defer the proof of \thmref{thm:maximization_bound} to the next subsection (\secref{sec:proof_thm_max}). In the following we show that an additive term is necessary in the lower bound. That is, we cannot remove the additive term (for example, we cannot push it into the multiplicative term involving $1-e^{-\ell/k}$).

Suppose the hidden state take two values $y_1, y_2$ and there are two test $t_1, t_2$. Let $\eta=0.1$. The conditional probabilities for the test outcomes are as follows: $p(t_1=1\given y_1) = p(t_1=1\given y_2) = 1$, $p(t_2=1 \given y_1) = 0.001, p(t_2=1\given y_2) = 0$. There are only two hypotheses with non-zero probability, i.e., $h_1 = (1,0)$ and $h_2 = (1,1)$. 
Further assume there are two distinct decisions $d_1, d_2$ that are optimal for hypotheses $h_1$ is $h_2$ respectively.

However, the sampler will output only one hypothesis $h_1 = (1,0)$, since $p(h_1 \given y_1) > 1-\eta$ and $p(h_2 \given y_2) > 1 - \eta$. Assume that we further add infinitely many ``dummy tests'', i.e., for all $t$ in this set, $p(t = 1\given y) = 0$ for all $y$. Then the greedy algorithm will choose those tests with high probability, since the gain for all tests over $\tH$ is 0; whereas a smarter algorithm will pick test $t_2$, because we can identify the target region (and hence obtain a positive gain) upon observing its outcome.

\subsection{PROOF OF \THMREF{thm:maximization_bound}}\label{sec:proof_thm_max}

Assume that the cumulative probability of the enumerate hypotheses is at least $1-\eta$, i.e., using our sampling algorithm we enumerate $1-\eta$ fraction of the total mass.

Denote the set of sampled hypotheses by $\tH$, and the expected gain of test $t$ on $\tH$ by $\Delta_{\tH}(t \given \cdot)$. Suppose we run the greedy algorithm based on $\tH$. We want to show that the following lemma holds:
\begin{lemma}\label{lm:perstepgain}
  Suppose $\tH \subseteq \cH$ and $p(\tH, \bx_\cA) \geq (1-\eta) p(\cH, \bx_\cA)$. Let $\tilde{t} \triangleq \argmax_t \Delta_{\tH}(t\given \bx_\cA)$ be the test with the maximal gain on $\tH$ in the \ECT objective function. Then for any test $t$, it holds that
  \begin{align*}
    \Delta_{\cH}(\tilde{t}\given \bx_\cA) \geq \Delta_{\cH}(t\given \bx_\cA) - 2\eta p(\bx_\cA)^2.
  \end{align*}
\end{lemma}
That is, the test $\tilde{t}$ which achieves the maximal gain on $\tH$ will achieve a gain on $\cH$ which is no less than $\varepsilon\triangleq 2\eta p(\bx_\cA)^2$ below the maximal gain of any test. In the following we provide the proof of Lemma~\ref{lm:perstepgain}.
\begin{proof}
  Clearly, if we can show that for any test $t$, the gain of $t$ over
  $\tH$ and the gain of $t$ over $\cH$ are at most $\varepsilon$
  apart, i.e.,
  \begin{align}
    \Delta_{\cH}(t\given \bx_\cA) \leq \Delta_{\tH}(t\given \bx_\cA) + \varepsilon \label{eq:diffpertest},
  \end{align}
  then we know that
  $ \Delta_{\cH}(t^* \given \bx_\cA) \leq \Delta_{\tH}(t^* \given \bx_\cA) + \varepsilon \leq \Delta_{\tH}(\tilde{t}\given \bx_\cA) + \varepsilon$.

  In the following, we show that inequality~\eqref{eq:diffpertest}
  holds.

  The conditional expected gain of test $t$ over observed tests
  $\bx_\cA$ is
  \begin{align*}
    \Delta_{\tH}(t \given \bx_\cA)
    &= \expctover{}{ \delta_{\tH}(x_t \given \bx_\cA )} \\
    & = p(x_t=1 \given \bx_\cA) \delta_{\tH}(x_t=1 \given \bx_\cA)
      + p(x_t=0 \given \bx_\cA) \delta_{\tH}(x_t=0 \given \bx_\cA).
  \end{align*}
  Here $\delta_{\tH}(x_t \given \bx_\cA)$ denotes the conditional
  benefit of test $t$ if its outcome is realized as $x_t$. Note that we can
  compute the probability terms $p(x_t=1 \given \bx_\cA)$ and
  $p(x_t=0 \given \bx_\cA)$ \emph{exactly} from the CPT $\{\theta_{ij}\}_{n\times m}$
  via Bayesian update,
  i.e.,
  $ p(x_t \given \bx_{\cA}) = \sum_{y} p(x_t, y \given \bx_{\cA}) =
  \frac{\sum_{y} p(y) p(\bx_{\cA} \given y) p(x_t \given y) } {\sum_y
    p(y) p(\bx_{\cA} \given y)}$.
  What remains to be approximated is the gain for each specific
  realization. For \ECT object function, the gain of observing $x_t$
  over hypothesis set $\cH$ after having observed $\bx_\cA$ is
  \begin{align*}
    \delta_{\cH}(x_t \given \bx_\cA) &= \sum_{i> j} (p(\region_i, \bx_\cA) p(\region_j, \bx_\cA)
                                       - p(\region_i, \bx_\cA, x_t) p(\region_j, \bx_\cA, x_t)),
  \end{align*}
  where $\region_i$ represent the set of hypotheses in \emph{region} / \emph{equivalence
    class} $i$.

  We define short-hand notation $\gamma_i := p(\region_i\setminus \tilde{\region}_i, \bx_A)$, where $\tilde{\region}_i$ denotes the sampled hypotheses of the $i^{\text{th}}$ decision region.
  The difference in the gain of $t$ over $\cH$ and $\tH$ can be expressed as
  \begin{align*}
    & \delta_{\cH}(x_t \given \bx_\cA)-\delta_{\tH}(x_t \given \bx_\cA) \\
    &= \sum_{i> j}\left(p(\region_i, \bx_\cA) p(\region_j, \bx_\cA) - p(\tilde{\region}_i, \bx_\cA) p(\tilde{\region}_j, \bx_\cA)\right) -
    \\ & \hspace{1cm} \sum_{i> j} \left(p(\region_i, \bx_\cA, x_t) p(\region_j, \bx_\cA, x_t) - \right.
         \left. p(\tilde{\region}_i, \bx_\cA, x_t) p(\tilde{\region}_j, \bx_\cA, x_t) \right)\\
    & \leq \sum_{i> j} \left(p(\region_i, \bx_\cA) p(\region_j, \bx_\cA) - \right. \left. p(\tilde{\region}_i, \bx_\cA) p(\tilde{\region}_j, \bx_\cA)\right) \\
    &= \sum_{i> j} \left( \left(p(\tilde{\region}_i, \bx_\cA) + \gamma_i\right) \left( p(\tilde{\region}_j, \bx_\cA) + \gamma_j\right) - \right.
      \left. p(\tilde{\region}_i, \bx_\cA) p(\tilde{\region}_j, \bx_\cA)  \right)\\
    &= \sum_{i> j} \left( \gamma_i \left(\gamma_j + p(\tilde{\region}_j, \bx_\cA)\right) + \gamma_j p(\tilde{\region}_i, \bx_\cA)  \right) \\
    &= \sum_{i>j}\left( \gamma_i p(\region_j, \bx_\cA) + \gamma_j p(\tilde{\region}_i, \bx_\cA) \right) \\
    &\leq \sum_{i}\gamma_i \sum_{j}p(\region_j, \bx_\cA) + \sum_j\gamma_j \sum_i p(\tilde{\region}_i, \bx_\cA).
  \end{align*}
  By the definition of $\gamma_i$ we know that
  \begin{align*}
    \sum_{i}\gamma_i &= p\left( \bigcup_{i} \left(\region_i \setminus \tilde{\region}_i
                       \right), \bx_\cA \right)
         \stackrel{(a)}{=} p\left( \left(\bigcup_{i} \region_i \setminus
         \bigcup_{i} \tilde{\region}_i \right), \bx_\cA \right)
         = p\left(\cH\setminus \tH, \bx_\cA\right) \leq \eta p(\bx_\cA).
  \end{align*}
  Step (a) is because of the assumption that $\region_i$'s do not overlap. Hence,
  \begin{align}
    \Delta_{\cH} (t \given \bx_\cA) - \Delta_{\tH}(t \given \bx_\cA)
                &= \expctover{}{ \delta_{\tH}(x_t \given \bx_\cA )} \nonumber \\
                &\leq \eta p(\bx_\cA) \sum_{j}p(\region_j, \bx_\cA) + \eta p(\bx_\cA) \sum_i p(\tilde{\region}_i, \bx_\cA) \nonumber \\
                &\leq 2\eta p(\bx_\cA)^2. \label{eq:diffacrossh}
  \end{align}
  Combining Equation~\eqref{eq:diffpertest} and \eqref{eq:diffacrossh} we finish the proof.
\end{proof}

Next, we provide the proof of \thmref{thm:maximization_bound} using the \lemref{lm:perstepgain}.
\begin{proof}[Proof of \thmref{thm:maximization_bound}] The key of the proof is to bound the one-step gain of the policy $\pi^g_{\tH,[\ell]}$.
  \begin{align*}
    f_{avg}&(\pi^g_{\tH,[i+1]}) - f_{avg}(\pi^g_{\tH,[i]}) \\
           &\stackrel{\text{\lemref{lm:perstepgain}}}{\geq} \expct{\max_t(\Delta(t\given \bx_\cA)) - 2\eta} \\
           &\stackrel{\text{(a)}}{\geq} \expct{\frac{\Delta(\policy^*_{\cH,[k]}\given \bx_\cA)}{k} - 2\eta} \\
           &= \expct{\frac{f_{avg}(\policy^*_{\cH,[k]}@\pi^g_{\tH,[i]}) - f_{avg}(\pi^g_{\tH,[i]})}{k} - 2\eta} \\
           &\stackrel{\text{(b)}}{\geq} \expct{\frac{f_{avg}(\policy^*_{\cH,[k]}) - f_{avg}(\pi^g_{\tH,[i]})}{k} - 2\eta}.
  \end{align*}
  Here $\policy^*_{\cH,[k]}@\pi^g_{\tH,[i]}$ denotes the concatenated policy of $\policy^*_{\cH,[k]}$ and $\pi^g_{\tH, [i]}$ (i.e., we first run $\pi^g_{\tH,[i]}$, and then run $\policy^*_{\cH,[k]}$ from scratch, ignoring the observations made by $\pi^g_{\tH, [i]}$).

  The proof structure follows closely from the proof of Theorem A.10 in \cite{golovin2011adaptive}: Step (a) follows from from the adaptive submodularity of $f$, and step (b) is due to monotonicity of $f_{avg}$. Define $\Delta_i := f_{avg}(\policy^*_{\cH,[k]}) - f_{avg}(\pi^g_{\tH,[i]})$, from the above equation we get $\Delta_\ell \leq \left( 1 - \frac{1}{k} \right)^l \Delta_0 + \sum_{i=0}^l \left( 1 - \frac1k \right)^i$. Hence, $ f_{avg}\left( \policy^g_{\tH,[\ell]} \right) \geq \left( 1- e^{-\ell/k} \right)f_{avg}\left( \policy^*_{\cH,[k]} \right) - 2k\eta \left( 1-\left(\frac{1}{k}\right)^\ell \right)$.
\end{proof}

\glsaddallunused


\printglossary[
type=symbolslist,
style=notationlong]   


\end{document}